\documentclass[runningheads]{llncs}
\usepackage{graphicx}
\usepackage{amsmath,amssymb} % define this before the line numbering.
\DeclareMathOperator*{\argmin}{arg\,min}
\usepackage{algorithm}
\usepackage[noend]{algpseudocode}
\makeatother
\usepackage[utf8]{inputenc}
\usepackage{wrapfig,lipsum,booktabs}
\usepackage{color}
\usepackage[width=122mm,left=12mm,paperwidth=146mm,height=193mm,top=12mm,paperheight=217mm]{geometry}
\begin{document}
% \linenumbers
\pagestyle{headings}
\mainmatter
\def\ECCV18SubNumber{2163}  % Insert your submission number here

\title{Face hallucination using cascaded super-resolution and identity priors} % Replace with your title

%\titlerunning{Submitted for review to ECCV 2018}

\authorrunning{Grm, Dobrišek, Scheirer, Štruc}

\author{Klemen Grm$^1$, Simon Dobrišek$^1$, Walter J. Scheirer$^2$, Vitomir Štruc$^1$}
\institute{$^1$ University of Ljubljana, Faculty of Electrical Engineering \\
$^2$ University of Notre Dame, Department of Computer Science and Engineering}

%%%%%%%%%%%%%%%%%%%%%%%%%%%%%%%%%%%%%%%%%%%%%%%%%%%%%%%%%%%%%%%
% Additional info that needs to go in for the next version
% Date: 19 March, 2018
%
%
% - explain kXnXsX notation in caption
% - explain how test data was generated
% - add D subscript in overall loss notation
% - add a few equations and symbols to the text, maybe include f: x_lr->x_hr and so on
% - say that we use a Gaussian filter for the approximation of the local statistics with SSIM
% - include numerical results for SSIM loss in experiments - ablation study
% - in Fig. 2 add a block saying detail-computation in front of the recognition models
% - add progressive and cascaded methods into related work, .e.g., http://openaccess.thecvf.com/content_cvpr_2017/papers/Lai_Deep_Laplacian_Pyramid_CVPR_2017_paper.pdf
% - generate results on a couple of other datasets, e.g., Helen, etc.
% cite "Loss Functions for Image Restoration With Neural Networks"
%
%
%%%%%%%%%%%%%%%%%%%%%%%%%%%%%%%%%%%%%%%%%%%%%%%%%%%%%%%%%%%%%%%%
\maketitle

\vspace{-9mm}

\begin{figure}[h!]
\centering
\includegraphics[width=1\columnwidth]{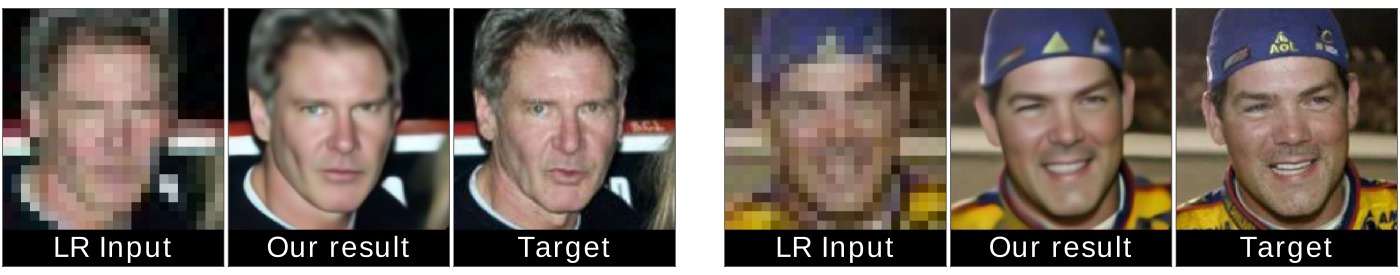}\vspace{-2mm}
\caption{Sample face hallucination results generated with the proposed method.}
\label{fig: teaserX}
\vspace{-10mm}
\end{figure}

\begin{abstract}
In this paper we address the problem of hallucinating high-resolution facial images from unaligned low-resolution inputs at high magnification factors. We approach the problem with convolutional neural networks (CNNs) and propose a novel (deep) face hallucination model that incorporates identity priors into the learning procedure. The model consists of two main parts: \textit{i)} a cascaded super-resolution network that upscales the low-resolution images, and \textit{ii)} an ensemble of face recognition models that act as identity priors for the super-resolution network during training. Different from competing super-resolution approaches that typically rely on a single model for upscaling (even with large magnification factors), our network uses a cascade of multiple SR models that progressively upscale the low-resolution images using steps of $2\times$. This characteristic allows us to apply supervision signals (target appearances) at different resolutions and incorporate identity constraints at multiple-scales. Our model is able to upscale (very) low-resolution images captured in unconstrained conditions and produce visually convincing results. We rigorously evaluate the proposed model on a large datasets of facial images and report superior performance compared to the state-of-the-art. 
%\keywords{Face hallucination, face recognition, super-resolution}
\end{abstract}

% The image triplets show: the input low-resolution images (left), the super-resolved outputs of our model (center), and ground truth images (right). Our method can super-resolve facial images of size $24 \times 24$ by a factor of $8\times$ in a manner that preserves identity information and resolves high-frequency details, even for images and subjects not present in the training set.

\section{Introduction}

Face hallucination represents a domain-specific super-resolution (SR) problem where the goal is to recover high-resolution (HR) face images from low-resolution (LR) inputs~\cite{baker2000hallucinating}. It has important applications in image enhancement, compression and face recognition~\cite{liu2007face}, but also surveillance and security~\cite{baker2002limits,gunturk2003eigenface}. 

Similar to other single-image super-resolution tasks, face hallucination is inherently ill-posed. Given a fixed image-degradation model, every LR facial image can be shown to have many possible  HR counterparts. Thus, the solution space for SR problems is extremely large and existing solutions commonly try to produce plausible reconstructions by "hallucinating" high-frequency information based on the provided LR evidence. While significant progress has been made in recent years in the area of super-resolution and face hallucination~\cite{yang2010image,kim2016accurate,yang2012single,timofte2013anchored,dong2014learning,Salvador2015naive,johnson2016perceptual,ledig2017photo,cao2017attention,xu2017learning,sajjadi2017enhancenet,tong2017image,Lai_2017_CVPR,Tai_2017_CVPR,Huang_2017_CVPR}, super-resolving arbitrary facial images, especially at high magnification factors, is still  an open and challenging problem, mainly due to:
\begin{itemize}
\item The ill-posed nature of the face hallucination problem, where the solution space is known to grow exponentially with an increase in the desired magnification factor~\cite{baker_pami2002}. Even with strong reconstruction constraints it is exceptionally difficult to find good solutions and devise methods that work well under a broad range of conditions. Even for domain-specific SR problems, such as face hallucination, where the solution space is constrained by facial appearances, there are still an overwhelming number of possible solutions. 
\item The difficulty of learning and integrating strong priors into the face hallucination models that sufficiently constrain the solution space beyond solely the visual quality of the reconstructions. Most of the existing priors utilized for super-resolution relate to specific image characteristics, such as gradient distribution~\cite{cho2012image}, total variation~\cite{wang2007fast}, smoothness~\cite{dai2007soft} and the like, and hence focus on the perceptual quality of the super-resolved results. If discernibility of the semantic content is the goal of the SR procedure, such priors may not be the most optimal choice, as they are not sufficiently task-oriented.
\end{itemize}

The outlined limitation are most evident for challenging face hallucination problems where tiny low-resolution images (e.g., $24\times 24$ pixels) of arbitrary characteristics need to be super-resolved at high magnification factors (e.g., $8\times$). In this paper, we try to address some of these limitations with a new hallucination model build around deep convolutional neural networks (CNNs). Our model, called C-SRIP, uses a Cascade of simple Super-Resolution models (referred to as SR modules hereafter) for image upscaling and Identity Priors in the form of pretrained recognition networks as  constraints for the training procedure. The SR models super-resolve the LR input images in magnification increments of $2\times$ and, consequently, allow for intermediate supervision at every scale. This intermediate supervision confines the explosion of the solution-space size and contributes towards more accurate hallucination results. To preserve identity-related features in the SR images, we incorporate pretrained recognition models into the training procedure, which act as identity constraints for the face hallucination problem. The recognition models are trained to respond only to the hallucinated high-frequency parts of the SR images and ensure that the added facial details are not only plausible, but as close to the true details as possible.  Due to availability of intermediate SR results, we incorporate the identity constraints at multiple scales in the C-SRIP model. Additionally, we introduce a novel loss function derived from the structural similarity index (SSIM, \cite{wang2004image}) that provides a stronger error signal for model training than the  loss functions commonly used in this area. 
 
%\subsection{Contributions}

Overall, we make three main contributions in this paper:
\begin{enumerate}
\item We propose a new CNN-based face hallucination model, C-SRIP, that integrates identity priors at multiple scales into the training procedure of a super-resolution network. To the best of our knowledge, this is the first attempt to exploit \textit{multi-scale identity information} to constrain the solution space of deep-learning based SR models. 
\item We introduce a \textit{cascaded SR network} architecture that super-resolves images in magnification steps of $2\times$ and offers a convenient and transparent way of incorporating supervision signals an multiple scale. Once trained, the SR network is able to hallucinate  tiny unaligned $24\times 24$ pixel LR images  at magnification factors of $8\times$ and produce realistic and visually convincing hallucination results as illustrated in Fig.~\ref{fig: teaserX}.    
\item We formulate a \textit{novel differentiable loss} function for SR models based on the concept of structural similarity (SSIM). The novel loss drives our SR model towards solutions of higher perceived quality, as it relates to a measure designed explicitly with the goal of modeling human image-quality perception.     
%\item
\end{enumerate}\vspace{-3mm}

\section{Related work}\label{Sec: related work}
In this section we discuss recent research on super-resolution and face hallucination with the goal of providing the necessary context for our work. For a more comprehensive coverage the reader is referred to the existing surveys on super-resolution and face hallucination, e.g.,~\cite{tian2011survey,nasrollahi2014super,wang2014comprehensive,nguyen2018super}.

\textbf{Super-resolution:} Recent solutions to the problem of single-image super-resolution (SR) are  dominated by learning-based methods that use pairs of corresponding HR and LR images to train machine learning models capable of predicting HR outputs given LR evidence~\cite{yang2010image,kim2016accurate,yang2012single,timofte2013anchored,dong2014learning,Salvador2015naive}. The learning procedures used with these models typically aim to minimize an objective function that quantifies the error between the ground truth HR images and the SR predictions. Common objectives in this area include the mean-squared-error (MSE), the mean-absolute-error (MAE) and other related error metrics. Our SR model follows the outlined learning paradigm, but different from existing SR methods, exploits a novel objective related to structural similarity (SSIM,~\cite{wang2003multiscale}), which  better models human image perception than simple pixel-based metrics, such as MSE or MAE. 

Our C-SRIP model is based on convolutional neural networks (CNNs) and in this sense is related to recent SR models that exploit CNNs for image upscaling, e.g.,~\cite{dong2014learning,kim2016accurate,johnson2016perceptual,ledig2017photo,xu2017learning,sajjadi2017enhancenet,tong2017image,Lai_2017_CVPR,Tai_2017_CVPR,Huang_2017_CVPR}. A common aspect of these models is that they super-resolve images in a single step and, while capable of producing impressive SR results, rely only on LR-HR image pairs for training. Our model, on the other hand, upscales the LR inputs in a cascaded manner and allows for supervision signals and constraints to be incorporated at multiple scales during training.  

Recent CNN-based SR models, e.g., \cite{kim2016accurate,ledig2017photo} exploit contemporary network architectures such as ResNets~\cite{he2016deep} and Generative Adversarial Networks (GANs, \cite{goodfellow2014generative}). These models are closely related to our work, as we also make heavy use of residual connections  and incorporate a generative and a discriminative network in our model. While we do not rely on GANs per se, our model does include a discriminative (classification) model that constrains the solution space of the generative SR network. However, our discriminative model is pre-trained and then frozen and not optimized alternatively with the generator, which greatly improves  training stability and still  results in realistic SR outputs.

Our work can also be seen as an extreme case of the perceptual-loss ($\ell_p$) image transformation model from~\cite{johnson2016perceptual}, which relies on comparisons of high-level features extracted from a pretrained secondary network as the learning objective for SR, instead of comparisons at the pixel level. Our model follows a similar idea, but uses identity (information a highest possible semantic level) to constrain the solution space of the generative SR network. Thus, instead of network features, our model considers the outputs of a pretrained network during training.

\textbf{Face hallucination and identity constraints:} Because the solution space of face hallucination models is typically constrained to a set of plausible facial appearances, remarkable performance has been achieved with hallucination models at much higher magnification factors than for general single-image SR tasks~\cite{yu2016ultra}. Similarly to other vision problems, the research is moving increasingly towards deep learning and considerable improvements have been achieved recently with CNN-based models, such as~\cite{yu2016ultra,jia2008generalized,jin2015robust,zhu2016deep,yang2013structured,zhou2015learning,farrugia2017face,yu2017face,yu2018imagining}. We contribute to this body of work in this paper with a novel deep face hallucination model. While the SR network of our model is general and applicable to arbitrary input images, we infuse domain-specific knowledge into the model through face recognition models. 

It needs to be noted that using identity information as a prior (or constraint) for SR models has been examined before~\cite{liu2005neighbor,li2009aligning}. Henning-Yeomans et al.~\cite{hennings2008simultaneous}, for example, formulated a joint optimization approach that maximized for super-resolution and face recognition performance simultaneously. This approach is conceptually similar to our work, but our approach is more general in the sense that it can be applied with any differentiable classification model. The approach from~\cite{hennings2008simultaneous} is focused only on linear feature extraction techniques, e.g., PCA~\cite{turk1991eigenfaces}.

Recent CNN-based face hallucination methods~\cite{yu2016ultra} have included secondary networks as constraints, which are trained jointly with the SR network. We found this to decreases training stability, so we instead use separately trained recognition and SR networks, where the former acts as a constraint for the latter.\vspace{-2mm} 

%Our C-SRIP model is related to other models that combine multiple CNNs and focus on discrimination and (input conditioned) image generation. A notable example of such models is the GAN-based face frontalization approach of Yin et al.~\cite{yin2017towards}. Here, the authors exploit a face recognition network to constrain a frontalization procedure. Our C-SRIP model is similar in spirit, but exploits multiple recognition models to constrain the SR network and relies on a different mechanism to incorporate the recognition constrains. A similar GAN-based model was presented by Tran et al. in~\cite{tran2017disentangled}. The model uses recognition constrains to combine multiple unconstrained facial images and synthesize a novel image of the subject under arbitrary poses. The model is again related to C-SRIP through the use of recognition constrains but is oriented toward facial reposing and not super-resolution.   

\section{Proposed method}\label{Sec: our model}

%In this section, we describe the proposed face hallucination method. First, we introduce a simple image degradation model that we use to motivate our face hallucination approach and show the necessity of appropriate priors and constraints. We then describe the structure of our model followed by a description of the structural similarity loss function and the identity prior exploited during model training. 
%\subsection{C-SRIP overview}

\begin{figure}[t!]
\centering
\includegraphics[width=0.95\columnwidth]{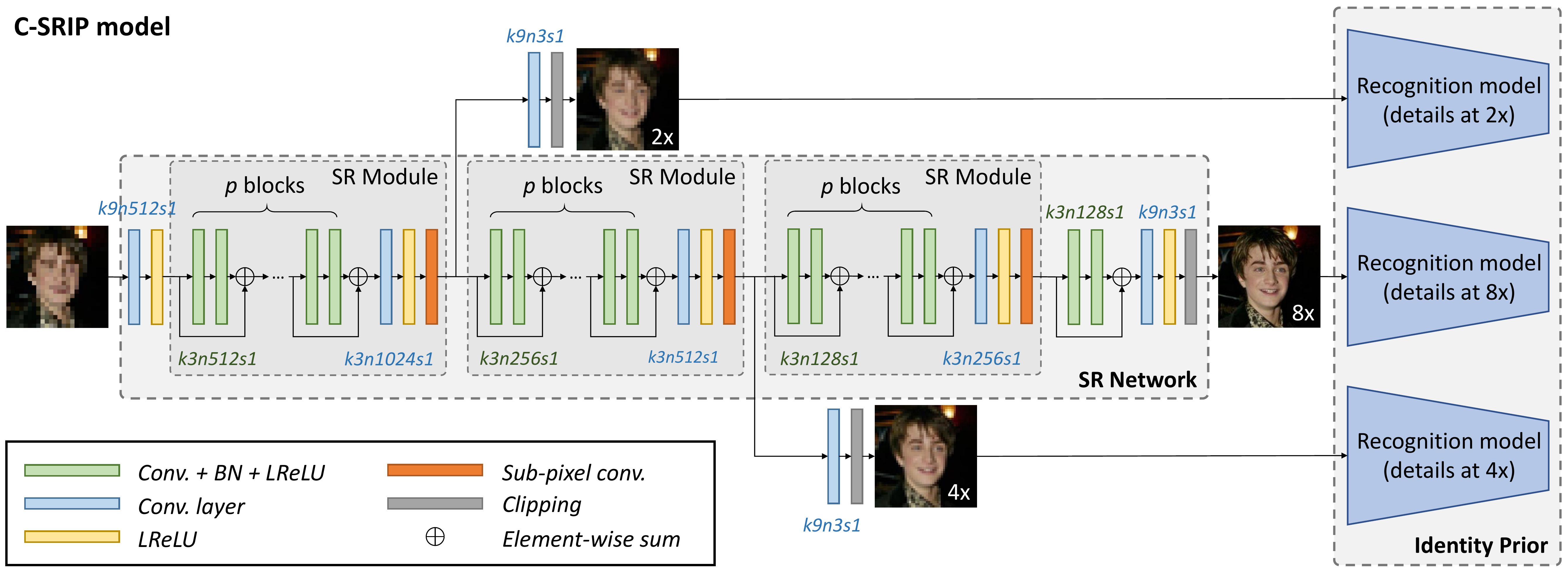}\vspace{-2mm}
\caption{Illustration of the proposed C-SRIP model. The model consists of a generative SR network and  an ensemble of face recognition models that serve as identity priors during training. The figure shows all architectural details (best viewed electronically).}
\label{fig: teaser}\vspace{-3mm}
\end{figure}

Our C-SRIP face hallucination model consists of two main components: \textit{i) a generative SR network} for image upscaling, build around a powerful cascaded residual architecture, and \textit{ii)} an ensemble of face recognition models that serve as identity priors for the C-SRIP model (see Fig.~\ref{fig: teaser}). In the following sections we describe all components of C-SRIP in detail and elaborate on the training procedure used to learn the model parameters. 
%Note that our C-SRIP model is the first to \textcolor{red}{Pohavlia se.} 

% of the input images that is    serve a  intermediate supervision aims to  a the supervision  in terms of reconstruction quality, but also identity discrimination. This    

\subsection{The cascaded SR network}\label{SubSec: SR model}

The generative part of our C-SRIP model is a $53$-layer deep convolutional neural network (CNN) that takes a LR facial image as input and super-resolves it at a magnification factor of $8\times$. The network progressively upscales the images using a cascaded series of so-called \textit{SR modules}. Each module upscales the image by a factor of $2\times$, which makes it possible to apply a loss function on the intermediate SR results and ensures better control of the training procedure in comparison to competing solutions that exploit supervision only at the final scale. The cascaded architecture allows us to solve a series of easier and better conditioned problems using repeated bottom-up inference with top-down supervision instead of one complex problem with an overwhelming amount of possible solutions. %makes it possible to solve a series  of  The overall effect of the cascaded architecture is that the model is required to   and contribution be used with supervision signals to  can be used at the output of every module contributing towards the intermediate outputs can be exploited to super-vise and allows for repeated bottom-up, top-down inference. For every upscaling step, we are able Since it is straight forward to produce ground-truth data for every scale between the LR input image and the target HR image, the     This approach allows the network to solve an easier problem   Instead of upscaling the image in a single step as    The generative part of C-SRIP, i.e., the SR network,  
%Our face hallucination model is a deep convolutional neural network. 

We design our SR  network around a fully-convolutional architecture that relies heavily on residual blocks~\cite{he2016deep} for all processing within one SR module and sub-pixel convolutions~\cite{shi2016real} for image upscaling. Our design choices are motivated by the success of fully-convolutional CNN models in various vision problems~\cite{he2016deep,krizhevsky2012imagenet,parkhi2015deep} and the state-of-the-art performance ensured by the sub-pixel convolutions in prior SR work~\cite{shi2016real,ledig2017photo}. Similarly to~\cite{ledig2017photo}, the residual blocks of the SR modules consist of two convolution--batch-norm--activation sub-blocks, followed by a post-activation element-wise sum. We ensure a constant memory footprint of all SR modules by decreasing the number of filters in the convolutional layers by a factor of $2$ with every upscaling step. This maximizes the capacity of the network and balances the computational complexity across the SR modules. %Thus, we use $512$ filters for all convolutional layers of the first SR module (at the initial scale), $256$ filters for the layers of the second SR module (at the $2\times$ scale), and $128$ filters for the last SR module and the final residual block at scale $8\times$. - SPECIFICS IN FIGURE
To upscale the feature maps at the output of each SR module, we rely on the sub-pixel convolution layers proposed in~\cite{shi2016real}. These layers increase the spatial dimensions of the feature maps  by reshuffling and aggregating pixels from multiple LR feature maps and, thus, for every upscaling step of $2\times$ reduce the number of available feature maps by a factor of $4\times$. We counteract this effect by doubling the number of filters in the convolutional layer preceding the sub-pixel convolutions and, consequently, ensure that the capacity of the SR modules is not compromised due to the upscaling. After reaching the target resolution, the feature maps are passed through one last residual block and a convolutional layer with $3$ output channels that produce the final $8\times$ super-resolved RGB image.   

The network branches off after each SR module to allow for intermediate top-down supervision during training. Each branch applies a series of large-filter convolutions to produce intermediate SR resolution results at different scales (i.e., $2\times$ and $4\times$ the initial scale) that are incorporated into the loss functions discussed in Section~\ref{Sec: loss}. However, these branches are not used at test time. The entire architecture of our network is illustrated in detail in Fig.~\ref{fig: teaser}.

\subsection{The identity prior}

Using prior information to constrain the solution space of SR models during training is a key mechanism in the area of super-resolution ~\cite{rudin1992nonlinear,wang2007fast,dai2007soft,shan2008high,lee2009geometric,sun2008image,cho2012image}. 
The main motivation for incorporating priors into SR models is to provide a source of additional information for the learning procedure that complements the commonly used reconstruction-oriented objectives and contributes towards sharper and more accurate SR results. 

An exceptionally strong prior in this context (also used in our model) is identity. Because identity information relates to the semantic content (i.e., who is in the image) and not the perceptual quality (i.e., how visually convincing is the image) of the SR images, it represents a natural choice for constraining the solution space of SR models. In fact, it seem intuitive to think about SR from both \textit{i) an image-enhancement} as well as a \textit{ii) content-preservation} perspective and to incorporate both views into the SR model %learning procedure 
for optimal results. While the image enhancement perspective is covered in our model by a reconstruction-based loss (discussed in Section \ref{Sec: loss}), the content-preservation aspect is addressed through an ensemble of  CNN-based face recognition models that ensure that identity information is not altered during upscaling.
\begin{figure}[t]
\begin{minipage}{0.58\textwidth}
\centering
\includegraphics[width=0.9\textwidth]{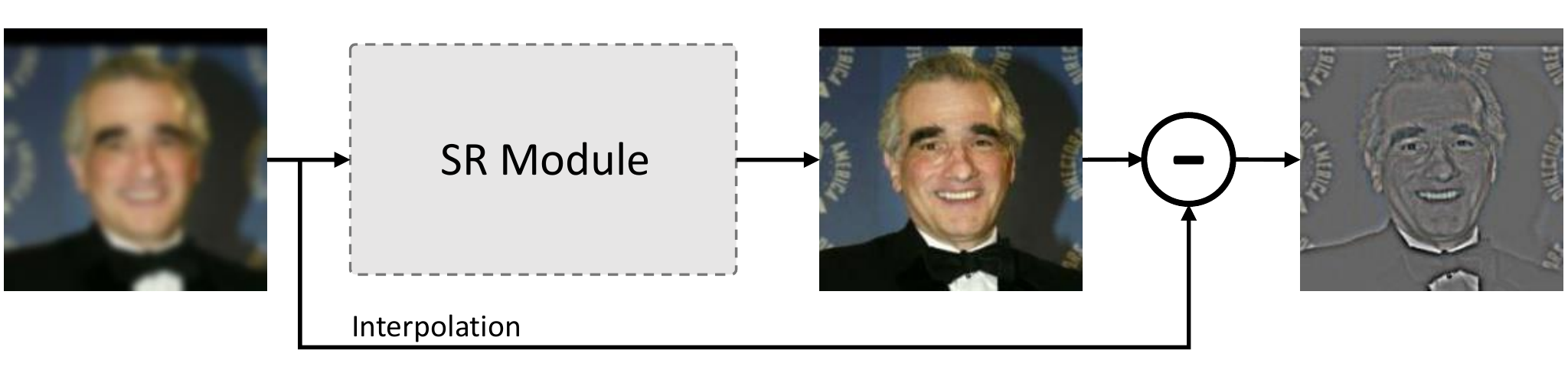}%,trim=1mm 0mm 0mm 0mm, clip
%\\
%\text{\footnotesize (a) Hallucinated details}
\end{minipage}
\hfill
\begin{minipage}{0.4\textwidth}
\centering
\includegraphics[width=0.9\textwidth]{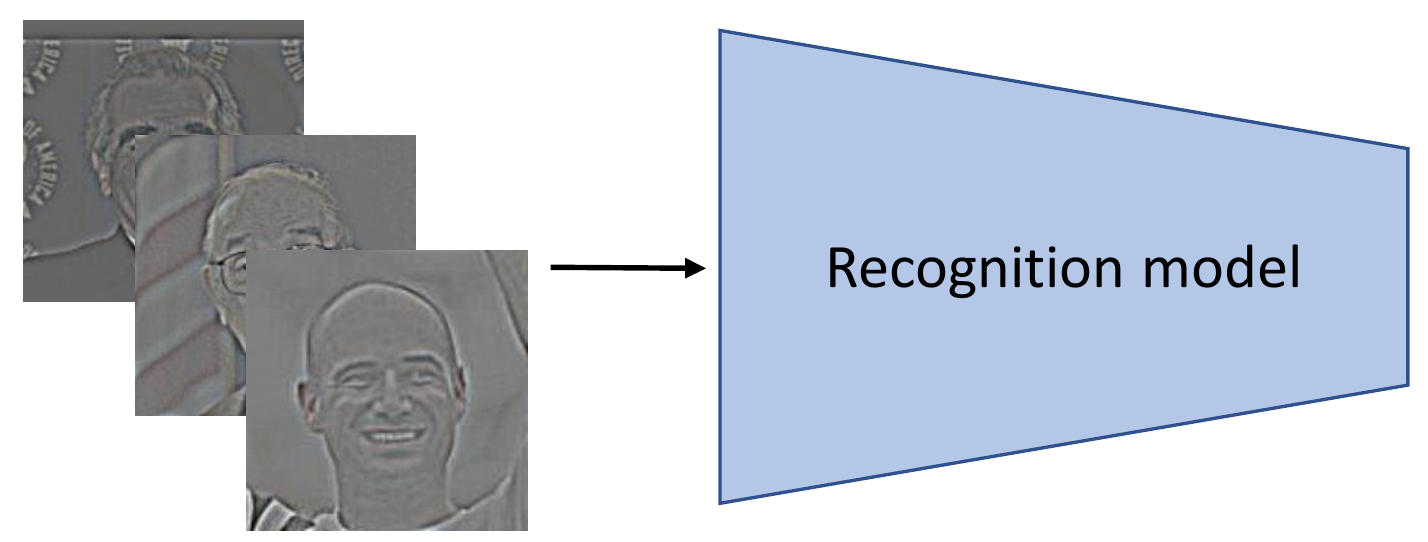}%,trim=1mm 0mm 0mm 0mm, clip
%\\
%\text{\footnotesize (b) Model input}
\end{minipage}
\caption{Each SR module adds fine facial details during upscaling (left). The recognition models are pre-trained to respond to these details only (right) and can therefore be used as identity constraints when learning the parameters of the SR network.}
\label{fig: rec_train}\vspace{-5mm}
\end{figure}

For C-SRIP we associate each recognition model with one of the SR modules and use it as an identity prior for the corresponding SR output, 
%The recognition models are included in C-SRIP . Each of the recognition models  is associated with one of the SR modules and serves as an identity prior for the corresponding SR output 
% i.e., for the hallucinated images at either $2\times$, $4\times$ or $8\times$ the initial image size, 
as illustrated in Fig.~\ref{fig: teaser}. Since each SR module can be shown to add only high-frequency details to the input images (see Fig.~\ref{fig: rec_train} left), we pretrain all recognition models to respond only to the hallucinated details and ignore the low-resolution content that is shared by  the input  and SR images (see Fig.~\ref{fig: rec_train} right). By focusing exclusively on the added details, we are able to directly link the recognition models to the desired SR outputs and penalize the results in case they alter the facial identity. This mechanism allows us to learn the parameters of the SR network by considering an identity-dependent loss in the  overall learning objective. 

While in principle any differentiable recognition model could be used as the identity prior for our face hallucination model, we select SqueezeNet models for this work~\cite{iandola2016squeezenet}. The main reason for our choice is the lightweight architecture of SqueezeNet, which does not impose significant runtime slowdowns due to its relatively small memory and FLOPS footprint. 

\begin{figure}[t]
\centering
\begin{minipage}{0.25\textwidth}
\centering
\includegraphics[height=0.9\textwidth]{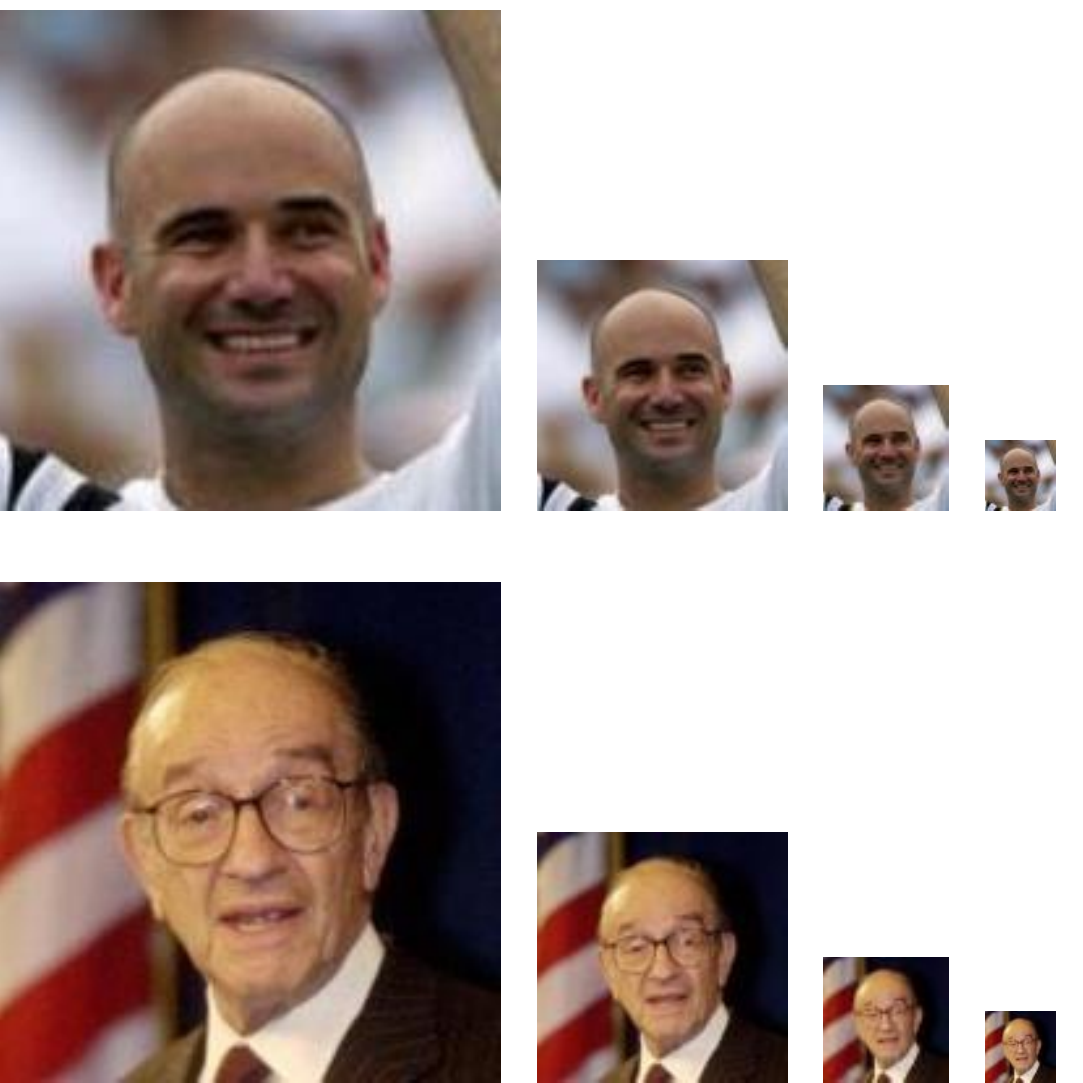}%,trim=1mm 0mm 0mm 0mm, clip
%\caption{(a)}
%\\
%\text{\footnotesize (a) Top down}
\end{minipage}
\hfill
\begin{minipage}{0.74\textwidth}
\centering
\includegraphics[height=0.31\textwidth]{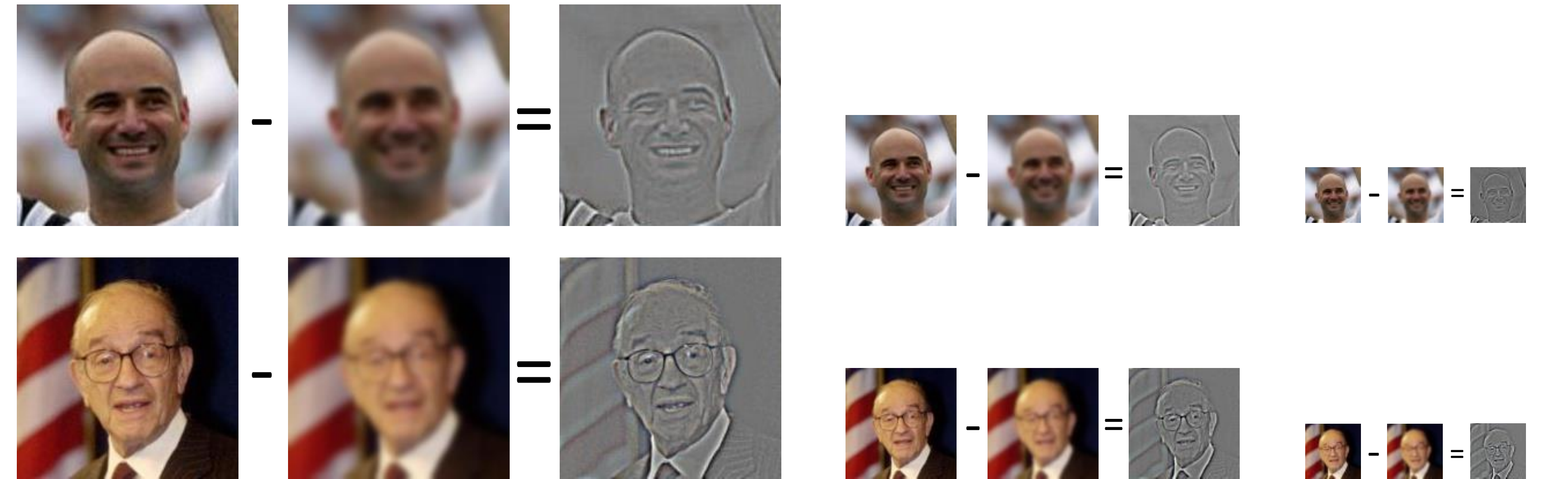}%,trim=1mm 0mm 0mm 0mm, clip
%\\
%\text{\footnotesize (b) Residual computation}
%\text{\footnotesize (a) Top down ($\sigma_n$ vs. $\sigma_b$)}
\end{minipage}
\caption{We generate training images for the SR network at four different spatial resolutions(left). For recognition-model training we compute residual images that correspond to the facial details that are hallucinated by the SR modules (right).}\vspace{-3mm}
\label{fig: residuals_training_data}
\end{figure}

\subsection{Training details and SSIM loss} \label{Sec: loss}

We train the C-SRIP model in two stages. In the first stage, we learn the parameters of the SqueezeNet models for all three SR outputs. In the second stage, we freeze the the weights of the recognition models and train the SR network with a combined loss. The details of both stages are presented next.

\textbf{Recognition-model training.} Next to LR and HR image pairs, we also require two intermediate reference images between the lowest and the highest resolution to learn the parameters of the recognition models and SR modules. To this end, we apply a simple degradation model on the  available  HR images $\mathbf{x}^{hr}_{i}$ and generate $N$ image quadruplets for training, i.e., $\{\mathbf{x}^{lr}_{i}, \mathbf{x}^{2\times}_i, \mathbf{x}^{4\times}_i, \mathbf{x}^{hr}_i \}_{i=1}^N$, where $\mathbf{x}^{lr}_{i}$ represents the LR input image, $\mathbf{x}^{2\times}_i$ and $\mathbf{x}^{4\times}_i$ stand for the intermediate SR results at $2\times$ and $4\times$ magnification factors, respectively, and the HR image  $\textbf{x}^{hr}_{i}$ corresponds to the ground truth for the magnification factor of $8\times$. 
Our degradation model uses Gaussian blurring followed by image decimation for down-sampling and produces training data as shown on the left side of Fig.~\ref{fig: residuals_training_data}. 

To train the recognition models, we construct residual images that reflect the facial details that need to be learned by the SR modules. The residual images, shown on the right side of Fig.~\ref{fig: residuals_training_data}, are computed by smoothing the ground truth images by a Gaussian kernel and subtracting the smoothed image from the original, i.e., $\Delta\mathbf{x}^j_i=\mathbf{x}^j_i-\mathbf{g}*\mathbf{x}^j_i$, for $j\in\{2\times, 4\times, hr\}$, where $\sigma$ values of $\sigma_{2\times}=1/3$, $\sigma_{2\times}=1$ and $\sigma_{8\times}=7/3$ are used with images at $2\times, 4\times$, and $8\times$ the LR image size, respectively.  We train the SqueezeNet models based on the generated residual images using categorical cross-entropy $L_{CE}$:\vspace{-1mm}
\begin{equation}
L_{CE}(\theta_{SN},\Delta\mathbf{x}) = -\sum_{k=1}^K p_{\Delta\mathbf{x}}(k)\,log\,\hat{p}_{\Delta\mathbf{x}}(k),\vspace{-0.5mm}
\end{equation}
%To avoid over-fitting we use data augmentation in the form of random horizontal flipping, random crops and small rotations ($\pm 5^\circ$). 
where $p_{\Delta\mathbf{x}}$ denotes the ground truth class probability distribution of the residual image ${\Delta\mathbf{x}}$ (i.e., $p_{\Delta\mathbf{x}}\in\{0,1\}^K$ is a class-encoded one-hot vector), $\hat{p}_{\Delta\mathbf{x}}\in\mathbb{R}^K$ stands for the output probability distribution produced by SqueezeNet's softmax layer based on ${\Delta\mathbf{x}}$, i.e.,  $K$ stands for the number of classes in the training data and $\theta_{SN}$ represents the parameters of the network. We learn the parameters of all three recognition models through backpropagation by minimizaing the $L_{CE}$ loss over the training dataset, i.e.:
$\hat{\theta}_{SN}^j=\argmin_{\theta_{SN}^j}\mathbb{E}_{\Delta\mathbf{x}^j}\left[L_{CE}(\theta_{SN}^j,\Delta\mathbf{x}^j)\right]$.
%is passed as an input to the network $f$ parametrized by weights $\theta$. $p_\mathbf{x}(c_i)$ refers to the probability of a specific class $c_i$ from the training set. 

The result of this first training stage are three SqueezeNet face recognition models $\hat{\theta}_{SN}^{2\times}, \hat{\theta}_{SN}^{4\times}$, $\hat{\theta}_{SN}^{hr}$, one for each image resolution  that respond only to the hallucinated facial details and serve as identity constraints for the SR network. 

\textbf{SR network training.} Standard reconstruction-oriented loss functions used for learning SR models, such as MSE or MAE, are known to produce overly smooth and often blurry SR results~\cite{ledig2017photo}. We therefore design a new loss function for our SR network around the  structural similarity index (SSIM, \cite{wang2003multiscale}), and integrate it directly into our learning algorithm. Specifically, we use our SSIM approximation as a loss function for the C-SRIP hallucination model. 

Given a ground truth image $\mathbf{x}$ and the corresponding SR network prediction $\hat{\mathbf{x}}=f_{\theta_{SR}}\left(\mathbf{x}\right)$, we compute the SSIM-based loss as follows:
\begin{equation}
%L_{SSIM}(\theta_{SR},\mathbf{x}) = \frac{1 - \mathbb{E}_x\left[\hat{\sc{SSIM}}(\mathbf{x},\hat{\mathbf{x}})\right]}{2},
L_{SSIM}(\theta_{SR},\mathbf{x}) = \frac{1}{2}\left(1 - \mathbb{E}_x\left[\hat{\sc{SSIM}}(\mathbf{x},\hat{\mathbf{x}})\right]\right),
\end{equation}
where  the SR network $f$ is parametrized by $\theta_{SR}$, $\mathbb{E}_x\left[\cdot\right]$ stands for the expectation operator over the spatial coordinates and $\hat{\sc{SSIM}}(\mathbf{x},\hat{\mathbf{x}})$ is a spatial similarity map between $\mathbf{x}$ and $\hat{\mathbf{x}}$ defined as: 
\begin{equation}
\hat{\sc{SSIM}}(\mathbf{x},\hat{\mathbf{x}})=\frac{\left(2\mu_{12} + C_1 \right ) \odot \left(2\sigma_{12} + C_2 \right )}{\left(\mu_1^2 + \mu_2^2 + C_1 \right) \odot \left(\sigma_1^2 + \sigma_2^2 + C_2 \right)}, \ \ \text{where}
\label{eq: modified SSIM}
\end{equation}
\begin{align*} 
\mu_1 &= \mathbf{x} \ast \mathbf{g}, \ \ \  &\mu_1^2 &= \mu_1 \odot \mu_1, \ \ \ &\sigma_{1}^2 = \left( \mathbf{x} \odot \mathbf{x}\right) \ast \mathbf{g} - \mu_1^2, \\
\mu_2 &= \hat{\mathbf{x}} \ast \mathbf{g}, \ \ \ &\mu_2^2 &= \mu_2 \odot \mu_2, \ \ \ & \sigma_{2}^2 = \left( \hat{\mathbf{x}} \odot \hat{\mathbf{x}}\right) \ast \mathbf{g} - \mu_2^2,\\
\mu_{12}  &= \mu_1 \odot \mu_2, \ \ \ &\sigma_{12} &= \left( \mathbf{x} \odot \hat{\mathbf{x}}\right)\ast \mathbf{g} - \mu_{12}.
\end{align*}
In the above equations, $\ast$ denotes the convolution operator, $\odot$ denotes the Hadamard product, and the open parameters, $\mathbf{g}$, $C_1$ and $C_2$, are defined as per the SSIM reference implementation provided by the authors of~\cite{wang2004image}, i.e., $\mathbf{g}$ is a $11\times 11$ Gaussian kernel with $\sigma = 1.5$ and  $C_1\approx6.55$, $C_2\approx58.98$. 

\begin{figure}[t!]
\centering
\includegraphics[width=0.99\columnwidth]{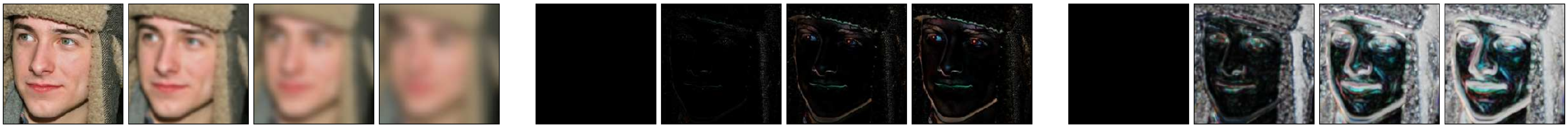}
\caption{Error maps generated by the squared-error used by the MSE loss and the proposed $\hat{\sc{SSIM}}$ function (the error map between $\mathbf{x}$ and the ground truth $\mathbf{x}_g$ is defined as $1-\hat{\sc{SSIM}}(\mathbf{x},\mathbf{x}_g)$). The figure represents degraded images (left), the corresponding squared-error maps (center) and the error maps generated by $\hat{\sc{SSIM}}$ (right).}\vspace{-3mm}
\label{mse_sslf_maps}
\end{figure}

In Fig~\ref{mse_sslf_maps}, we present error maps generated when comparing images of different resolutions with the ground truth based on squared-differences (center) and our $\hat{\sc{SSIM}}$ approximation (right). The examples show that the SSIM approximation  results in error maps that are less sparse compared to the squared-differences used with MSE-based losses, which, as we discuss in the experimental section, results in better training characteristics. 

Based on the pretrained SqueezeNet models and the loss introduced above, we defined the overall loss of our C-SRIP face hallucination model as follows:
\begin{equation}
{L}(\theta_{SR},\{\mathbf{x}^j\}) = \sum_{j\in\mathcal{D}}L_{SSIM}(\theta_{SR},\mathbf{x}^{j})+\alpha L_{CE}(\theta_{SN}^j,\Delta\mathbf{x}^j), 
\label{eq: overall loss}
\end{equation}
where $\mathcal{D}=\{2\times, 4\times, hr\}$, $\alpha$ is a weight parameter that balances the relative impact of the reconstruction- and recognition-based losses and $\theta_{SR}$ stands for the parameters of the SR network that we aim to learn. The residual images $\Delta\mathbf{x}^j$ are constructed during training as illustrated in Fig.~\ref{fig: residuals_training_data} (right). We use backpropagation to minimize the loss over our training data and find the parameters of the SR network $\hat{\theta}_{SR}$, i.e., $\hat{\theta}_{SR}=\argmin_{\theta_{SR}}\mathbb{E}_{\mathbf{x}^j}\left[L(\theta_{SR},\{\mathbf{x}^j\})\right]$. 

%our model through Our learning objective Note that ,
%To assist in learning image reconstruction at a very high magnification factor, our network doesn't only output the full $8\times$ magnified image, but also has output layers at the $2\times$ and $4\times$ scale. We use all three outputs for supervised learning. The entire loss function used for training the face hallucination network is \ref{full_loss},

%\subsection{Face hallucination with C-SRIP}

Once the training is complete, we remove the recognition models and network branches used to generate the intermediate SR results at $2\times$ and $4\times$ magnification factors and use only the main output of the SR network for face hallucination. The final SR network takes a LR image $\mathbf{x}_{lr}$ of size $24\times 24$ pixels as input and returns an $8\times$ upscaled $192\times 192$ facial image $\mathbf{x}_{hr}$ at the output. %We also remove the network branches used to generate the intermediate SR results at $2\times$ and $4\times$ the LR image size at test time. %, however, these could be used to predict additional SR results at smaller magnification factors if needed.  

%As we show in the experimental section, our SR model ensures state-of-the-art hallucination results for very low-resolution input images (i.e., $24\times 24$ pixels) and large upscaling factors (i.e., $8\times$). It makes no assumptions with respect to the characteristics of the facial images and is able to process unaligned LR inputs captured in  unconstrained settings.       

\subsection{Implementational details}

\textbf{Recognition models.} All three SqueezeNet models are implemented in accordance with the so-called \textit{complex SqueezeNet} architecture from~\cite{iandola2016squeezenet}. The models consist of $9$ fire modules with intermediate shortcut connections, followed by a global average pooling layer and a softmax classifier on top. We train the first recognition model to classify residual images at $2\times$ the initial LR scale, i.e., $48\times 48$ pixels, the second to classify images at $4\times$ the initial scale, i.e., $96\times 96$ pixels, and the last for recognition of residual images of $192\times 192$ pixels in size. To learn the model parameters we use backpropagation and the Adam~\cite{kingma2014adam} minibatch gradient descent algorithm, with a batch size of $128$ and an initial learning rate of $10^{-4}$. The learning rate is multiplied by a factor of $\frac{1}{3}$ every 20 epochs. To avoid over-fitting, we resort to data augmentation in the form of random horizontal flipping and random crops. We employ an early stopping criterion based on accuracy improvements on the validation set. If no improvements are observed over $10$ consecutive training epochs we stop the learning procedure and assume the recognition model has converged. 

\textbf{The SR network.} The SR network consist of three SR modules that are preceded by a  convolutional layer with $512$ large-scale filters of size $9\times 9$ pixels. The SR modules are implemented with $p=7$ residual blocks that contain $512$ filters in the first SR module, $256$ filters in the second SR module, and $128$ filters in the last SR module, as shown in Fig.~\ref{fig: teaser}. We set the number of filters for the final convolutional layer of the SR modules, to $1024$ for the first, $512$ for the second and $256$ for the third module. All filters are of size $3\times 3$ pixels. For the activations, we use Leaky Rectified Linear Units (LReLU). The last residual block of the SR network has $128$ filters $3\times 3$ pixels in size. Before generating SR results at the output of the network and in the off-branches, a convolutional layer with three $9\times 9$ filters is used followed by a clipping layer to ensure that the SR RGB images are within the valid intensity range of $\left[0,255\right]$.

%Each full-resolution image is blurred using a Gaussian kernel with $\sigma_b= \frac{r - 1}{3}$, where $r$ is the down-sampling ratio, e.g. $8$ when down-sampling images of size $192 \times 192$ to $24\times 24$. The blurred image is then decimated to the target size using bilinear interpolation.
%We first train the SqueezeNet network for from scratch for identity classification on the CASIA WebFace~\cite{yi2014learning} dataset using the categorical cross-entropy loss function. We chose this dataset because it features a wide variation in image quality, and was explicitly designed to include zero overlap with the LFW~\cite{huang2007labeled} dataset we used for evaluation. We use data augmentation in the form of random horizontal flipping and random crops at scales of between $180$ and $224$ pixels in size, and the images are additionally pre-processed by extracting the high-frequency residuals by subtracting a Gaussian-blurred image. Once the SqueezeNet network is trained to convergence, we freeze its weights and train the combined network to minimize the loss function described in \ref{full_loss} using the Adam optimization method~\cite{kingma2014adam} with a linear learning rate annealing strategy between $10^{-4}$ and $10^{-6}$ over the full training run of 100 epochs of $2^{16}$ images sampled from the dataset from a uniform distribution over the subject identities and downsampled at ratios of $8\times$, $4\times$ and $2\times$ to produce training input-output pairs for all three outputs of our face hallucination network.

We train the SR network based on the objective in Eq.~\eqref{eq: overall loss} that considers the novel SSIM-based loss as well as the recognition performance of the SqueezeNet models. We keep the parameters of the recognition models fixed and learn only the parameters of the SR network of C-SRIP with a value of $\alpha=0.001$. We again backbropagation and the Adam~\cite{kingma2014adam} minibatch gradient descent algorithm for training. Due to the large memory footprint of the SR network and the face recognition models, we use a relatively small batch size of $8$. We set the initial learning rate to $\frac{10}{3}\times 10^{-3}$ and multiply it by $\frac{1}{3}$ at the end of epochs $10$, $25$, $50$ and $80$. We use a combined early stopping criterion that assumes the model has  converged if both SSIM and MSE show no improvements over $10$ epochs.

%\begin{table}[!tb]
%\centering
%\caption{Recognition accuracy of the trained SqueezeNet models reported in the form of rank one recognition rates. The models are trained with  residual high-frequency images (i.e., images of the facial details) of different resolutions.}\label{tab: recognition model performance}
%\small
%\begin{tabular}{|c|c|c|c|}\hline
%Data & SqueezeNet ($48\times 48$) & SqueezeNet ($96\times 96$) & SqueezeNet ($192\times 192$) \\ \hline \hline
%Training data   & $0.5138$ & $0.7215$ & $0.8569$ \\  \hline
%Validation data & $0.2974$ & $0.4266$ & $0.5713$ \\  \hline
%\end{tabular}
%\end{table}

%\begin{table}[!tb]
%\centering
%\caption{\textcolor{red}{training set MSE and SSLF reached by simple generators}.}\label{tab: simple_generators}
%\small
%\begin{tabular}{|c|c|c|}\hline
%Data & MSE-trained model & SSLF-trained model \\ \hline \hline
%MSE   & $95.5719$ & $0.0811$ \\  \hline
%SSLF  & $81.4337$ & $0.0675$ \\  \hline
%\end{tabular}
%\end{table}

\section{Experiments}\label{Sec: experiments}

%In this section, we outline the experiments performed to validate the performance of our model. 

\subsection{Datasets and model training}

We select two datasets for our experiments. To train the C-SRIP model we use the CASIA WebFace dataset~\cite{yi2014learning} which features $494,414$ images of $10,575$ identities, (i.e., $N=494,414$; $K=10,575$). The CASIA WebFace images are blurred and sub-sampled to produce the necessary image quadruplets for training and employed for learning the parameters of the recognition models and the SR network (see Fig.~\ref{fig: residuals_training_data} for an illustration of the training-data generation process). For testing, we use the Labeled Faces in the Wild (LFW)~\cite{huang2007labeled} dataset with $13,233$ facial images and $5,749$ subjects. The two datasets are selected for the experiments because they feature images of variable quality captured in  unconstrained conditions and thus represent a significant challenge for SR models. More importantly, they are designed to contains zero overlap in terms of identity, which is paramount to ensure a fair and unbiased evaluation of the C-SRIP model. % as the identities in our test set have not been used to train the recognition model

For SqueezeNet training we randomly sample identities from CASIA WebFace and utilize $90\%$ of the images for training and $10\%$ for validation. The recognition models converge to the rank one recognition rate of $0.5138$ ($0.2974^\dagger$) with $48\times 48$px images, $0.7215$ ($0.4266^\dagger$) with $96\times 96$px images and $0.8569$ ($0.5713^\dagger$) with $192\times 192$px residual images on the training ($^\dagger$validation) data. As expected, the performance decreases with a decreasing size of the residual images and is adversely affected by the lack of low-frequency information during training (see, e.g., \cite{grm2017strengths} for the expected performance of SqueezeNet for face recognition). Nevertheless, the models contribute towards  accurate and visually  convincing SR results, as evidenced by the results in the next sections.   Since we also need identity information when learning the parameters of the SR network of C-SRIP, we again use the $90\%$/$10\%$ data split per identity for training and validation. With this setup we train the SR network on  $494,414$ CASIA WebFace images.  
\begin{figure*}[t!]
\text{\scriptsize \hspace{0.3mm}LR Input \hspace{0.35mm} Bicubic \hspace{0.45mm} NBSRF \hspace{1mm} SRCNN \hspace{1.8mm} VDSR \hspace{4.7mm} $\ell_p$ \hspace{4.4mm} SRGAN \hspace{0.05mm} URDGN \hspace{0.5mm} C-SRIP \hspace{1.3mm} Target \hfill}
{\centering
\includegraphics[width=1\textwidth]{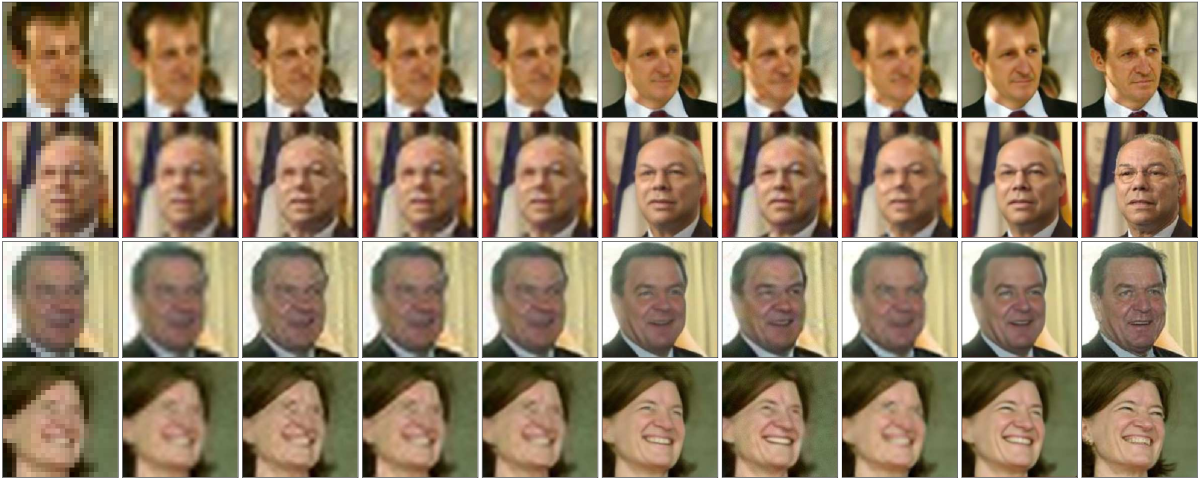}}\vspace{-2mm}
\caption{Qualitative comparison of state-of-the-art SR models on sample images from the LFW dataset. The first column shows the input $24\times 24$ pixel LR image (upscaled with nearest neighbor interpolation). Best viewed zoomed in.}\vspace{-3mm}
\label{lfw_restoration_fig}
\end{figure*}

We train all models on a workstation with two Nvidia GTX Titan Xp GPUs. On this hardware, the SqueezeNet training takes 1, 2, and 5 days, respectively, for the $2\times$, $4\times$ and $8\times$ scale models. The training of the SR network with the identity constraints included takes around 8 days. Once trained, the SR network is capable of processing images at an average speed of $15$~ms/image on GPU in batch mode, or $30$~ms/image in real-time (i.e., single-sample batch) mode. \vspace{-3mm}

\subsection{Comparison to the state-of-the-art}%comparison to the state of the art

We compare our C-SRIP model with $6$ state-of-the-art SR and face hallucination models, i.e.:
the Naive Bayes Super-Resolution Forest (NBSRF) from \cite{Salvador2015naive}, the Super-Resolution Convolutional Neural Network (SRCNN) from \cite{dong2014learning}, the Very Deep Super Resolution Network (VDSR) from \cite{kim2016accurate}, the perceptual-loss based SR model ($\ell_p$) from \cite{johnson2016perceptual}, the Super-Resolution Generative Adversarial Network from \cite{ledig2017photo}, and the Ultra Resolving Discriminative Generative Network (URDGN) from \cite{yu2016ultra}.   We train all models with the same data as C-SRIP and use open-source implementations of the authors (where available) for a fair comparison. For $\ell_p$ we use features from the fire2, fire3 and fire4 layers of SqueezeNet for the learning criterion. We include results for bicubic interpolation as a baseline. 

\textbf{Qualitative comparison.} A few sample SR images are presented in Fig.~\ref{lfw_restoration_fig}. We see that with magnification factors of $8\times$, interpolation methods are insufficient and result in the loss of facial details. Furthermore, general SR models, such as NBSRF, SRCNN and VDSR, fail to provide substantial improvements and are seen to amplify noise present in the LR images. These models fail to make use of the available facial context due to their relatively low receptive fields. The SRGAN, URDGN and $\ell_p$ models improve on this by including secondary networks as constraints during SR training. $\ell_p$ is consistently the best-performing model included in our comparison, only slightly behind C-SRIP. However, we notice it often adds high-frequency noise  when trying to minimize the perceptual loss of the convolutional maps of the secondary network. %This happens despite our best attempts to balance reconstruction- and perceptual-loss terms during  training. 
We speculate the reason our model is not susceptible to these errors is the global cross-entropy loss of the secondary networks as opposed to the local conv features exploited by $\ell_p$.
\begin{figure*}[t!]
\centering
\vspace{-4mm}
\includegraphics[width=0.9\textwidth]{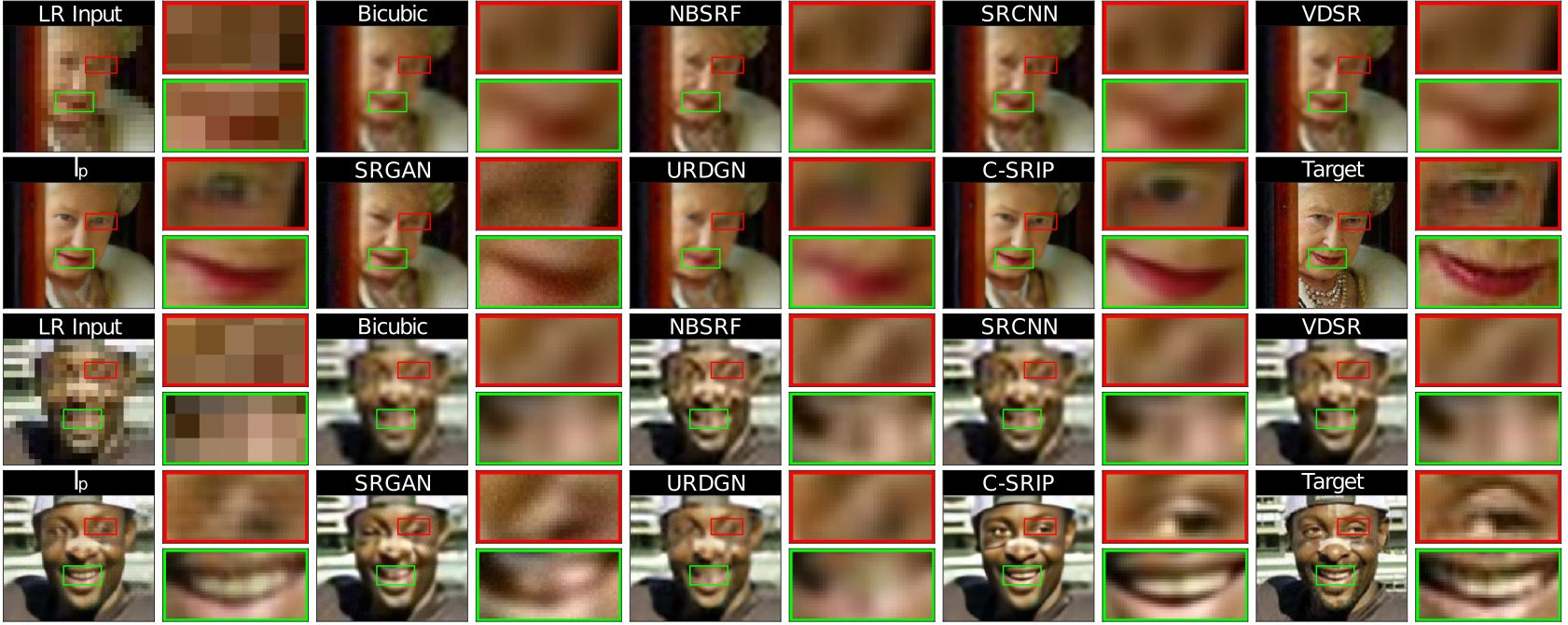}\vspace{-2mm}
\caption{Qualitative comparison of the evaluated SR models on sample images from the LFW dataset with highlighted image details. Best viewed electronically.}
\label{lfw_restoration_fig_insets}\vspace{-1mm}
\end{figure*}

\begin{table}[!tb]
\centering
\caption{Averaged PSNR and SSIM scores for the tested SR models computed over the LFW dataset. The  highest PSNR and SSIM values are achieved by  C-SRIP.}
\footnotesize
\label{lfw_restoration_table}
\renewcommand{\arraystretch}{1.1}
\resizebox{\columnwidth}{!}{%
\begin{tabular}{|l||c|c|c|c|c|c|c|c|} \hline
 Model & Bicubic & NBSRF~\cite{Salvador2015naive}    & SRCNN~\cite{dong2014learning}    & VDSR~\cite{kim2016accurate}    & \hspace{3mm}$\ell_p$~\cite{johnson2016perceptual} \hspace{3mm}& SRGAN~\cite{ledig2017photo} & URDGN~\cite{yu2016ultra} & C-SRIP (ours)  %\\
%        &         &  (ICCV,~\cite{Salvador2015naive}) & (PAMI, & (cvpr'16) &  (eccv'16) & (cvpr'17) & (eccv'16) &   (ours)        
\\\hline \hline
 PSNR & $24.256$ & $25.092$ & $24.812$ & $25.415$ & $26.985$ & $25.669$ &  $25.575$ & $\bf{27.164}$ \\
 SSIM & $0.7060$ & $0.7268$ & $0.7187$ & $0.7411$ & $0.7903$ & $0.6993$ & $0.7516$ & $ \bf{0.8171}$ \\ \hline
\end{tabular}}\vspace{-5mm}
\end{table}

\textbf{Quantitative comparison.} We report  average peak-signal-to-noise-ratio (PSNR) and structural similarity (SSIM) scores computed over the LFW images for all tested models in Table~\ref{lfw_restoration_table}. C-SRIP results in the best overall performance in terms of PSNR and SSIM, followed by $\ell_p$ and URDGN. While providing reasonably convincing visual results, SRGAN produces only an average PSNR score and the lowest SSIM score among all tested models. This result is expected and is observed regularly in the literature~\cite{ledig2017photo} with GAN-based SR methods. NBSRF, SRCNN and VDSR improve upon the Bicubic baseline in terms of performance metrics, but are less competitive in comparison to the three top performers of our experiments. 

\begin{figure}[tb]
    	\centering
        \includegraphics[width=0.95\linewidth]{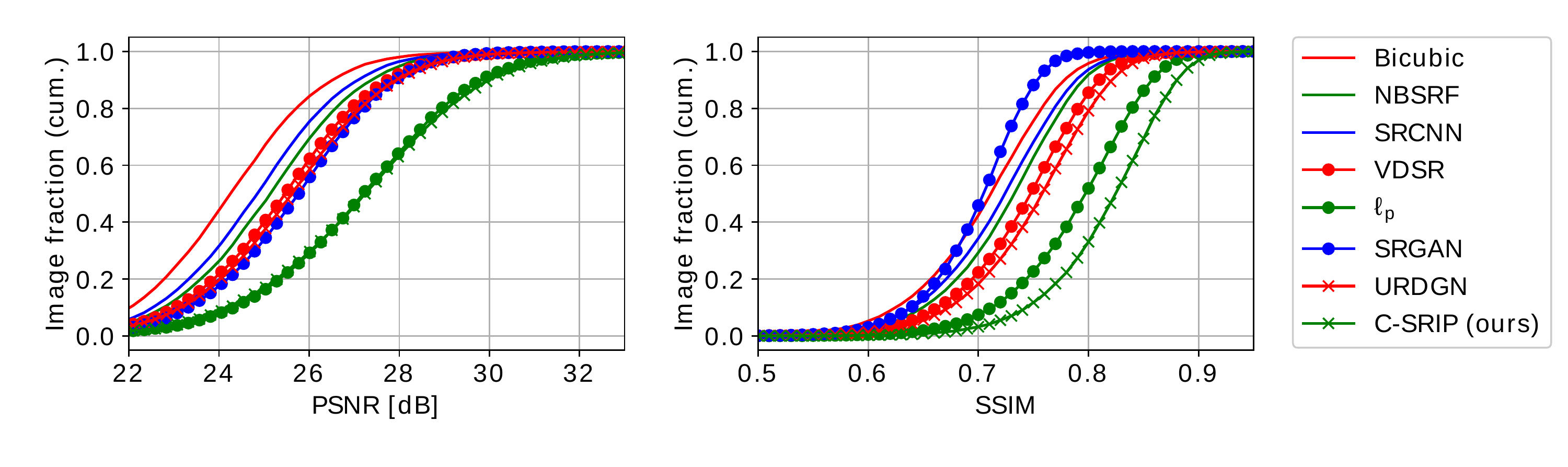}  \vspace{-6mm}      
		\caption{Cumulative Score Distribution curves (CSD) for the PSNR (left) and SSIM (right) scores over the LFW dataset. Curves further to the right are better.}
		\label{methods_cumdist}\vspace{-3mm}
	\end{figure}

The summary statistics in Table~\ref{lfw_restoration_table} show a partial picture of the performance of the tested models. To get better insight into the performance we present Cumulative Score (PSNR and SSIM) Distribution (CSD) curves of the experiments in Fig.~\ref{methods_cumdist}. Since SR models are increasingly  focusing on learning-based techniques, which are expected to perform inconsistently across images of different characteristics, CSD curves provide a reasonable way of visualizing this performance variability. From the presented curves we see that all tested methods vary significantly in PSNR and SSIM scores across the LFW dataset, with a large fraction of images producing sub-average performance scores. The $\ell_p$ and the proposed C-SRIP models are superior to other models and very close in terms of the PSNR-based CSD curve. However, the difference becomes significantly larger with the SSIM-based CSD curve, where C-SRIP is the  top performer. %The remaining models are less competitive and produce weaker CSD curves. %Similar to the    

%We believe CSD plots are better suited for comparing learning-based SR models than summary statistics, as 

 %We perform experiments on artificially degraded images from the LFW~\cite{huang2007labeled} dataset to evaluate the quality of our model's image restoration capabilities as well as its utility for face recognition.

%We first evaluated our choice of loss function and identity prior by performing an image restoration experiment on the LFW~\cite{huang2007labeled} dataset. We degraded all images using Gaussian blurring and $8\times$ downsampling. We present a sample of the results in figure \ref{lfw_restoration_fig}, and we present the image restoration metrics over the entire dataset in table \ref{lfw_restoration_table}.

\subsection{Ablation study}

We perform an ablation study with the goal of assessing the contribution of the individual components of our proposed C-SRIP model. Towards this end, we train the following models and evaluate their performance on the LFW dataset:
\begin{enumerate}
	\item {\it Baseline}: A baseline SR model without the cascaded SR modules. The model consist of 21 residual blocks similarly to our C-SRIP model, but the three sub-pixel convolution layers for upscaling are all at the end of the model. The model is trained using standard MSE loss.
    \item {\it B+SSIM}: Same as above, but trained with the proposed SSIM-based loss.
    \item {\it C+SSIM}: Our cascaded SR model, trained with the proposed SSIM-based loss, but without the identity prior networks and without multi-scale supervision i.e., the loss function is only applied at the output of the model. 
    \item {\it C+SSIM+M}: Our cascaded SR model, trained with multi-scale supervision and the proposed SSIM-based loss function, but without the identity priors.
    \item {\it C-SRIP}: The C-SRIP model with multi-scale SSIM and identity supervision.
\end{enumerate}

\begin{figure*}[t!]
{\centering
\includegraphics[width=1\textwidth]{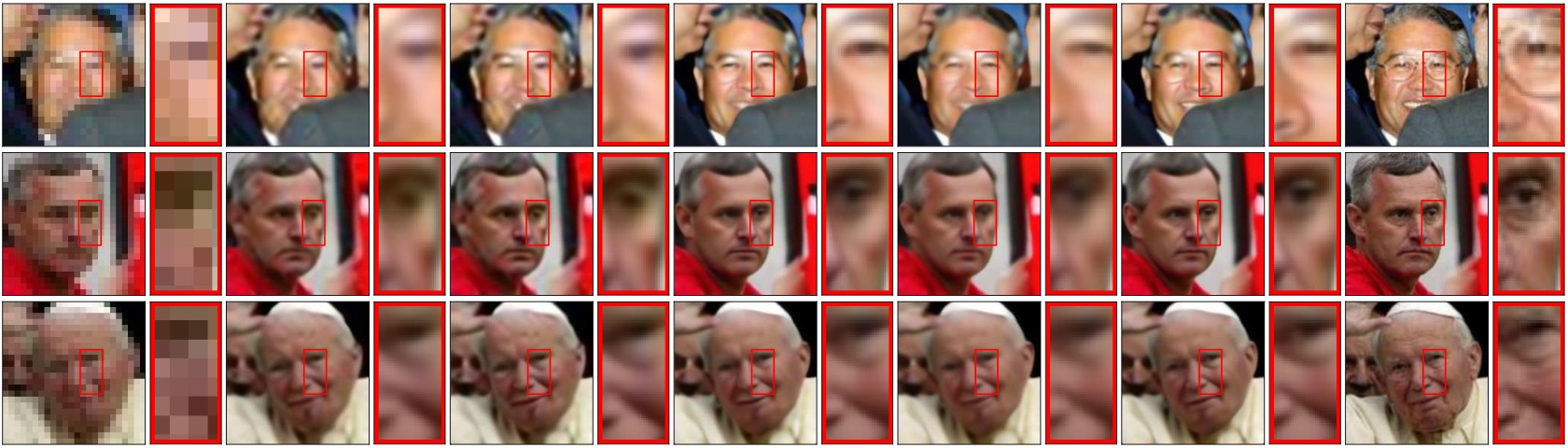}}
\text{\scriptsize \hspace{2.5mm} LR Input \hspace{4mm}  Baseline \hspace{5mm}  B+SSIM  \hspace{5mm} C+SSIM  \hspace{2.5mm} C+SSIM+M  \hspace{3.5mm} C-SRIP \hspace{7mm} Target}\vspace{-2mm}
\caption{Visual result of the ablation study.}
\label{ablation_study_fig_insets}\vspace{-1mm}
\end{figure*}

The results of the ablation study in Table~\ref{ablation_study} and the corresponding sample images in Fig.~\ref{ablation_study_fig_insets} show that each added component improves performance. The only decrease we see is when we switch from the MSE loss to the SSIM-based loss, which slightly lowers the average PSNR score, but results in a higher SSIM score. This result is expected, as PSNR is directly proportional to MSE and, thus, SR models optimizing for MSE typically achieve lower PSNR values than models using other loss functions. Nevertheless, we observe much better training characteristics with the SSIM loss, since the models converged faster and achieved significantly better SSIM and MSE scores on the training and validation data than the MSE-based models. Among the evaluated components, we see the biggest increase in the PSNR and SSIM scores with the multi-scale identity supervision. This addition also results in the biggest visual improvement of the SR images as seen in Fig.~\ref{ablation_study_fig_insets}.
\begin{table}[tb]
\centering
\caption{Ablation study on the LFW dataset. The table shows the impact of different model components on the average PSNR and SSIM scores.}
\small
\footnotesize
\label{ablation_study}
\begin{tabular}{|l|r|r|r|r|r|} \hline
 & Baseline & B+SSIM & C+SSIM & C+SSIM+M & C-SRIP\\ \hline \hline
 PSNR & $26.1748$ & $26.0251$ & $26.4137$ & $26.4511$ & $27.1638$ \\
 SSIM & $0.7547$ & $0.7579$ & $0.7731$ & $0.7841$ & $0.8171$ \\ \hline
\end{tabular}\vspace{-3mm}
\end{table}

\subsection{Limitations of C-SRIP}

\begin{figure*}[t!]
\centering
\includegraphics[width=0.9\textwidth]{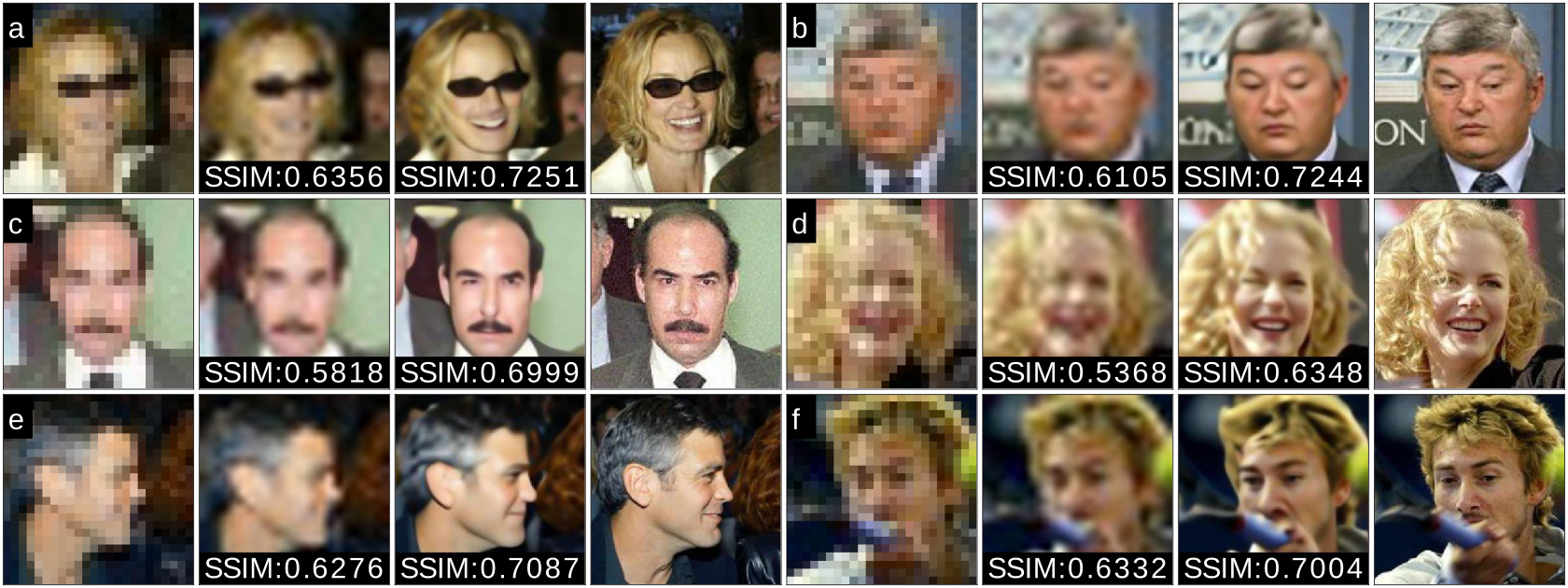}\vspace{-2mm}
\caption{Examples of poor SR results obtained with the C-SRIP model according to the SSIM value. The four columns of each image correspond to (from left to right): the input LR image, bicubic interpolation, C-SRIP and the target HR image.}\vspace{-2mm}
\label{failure_cases}
\end{figure*}

To evaluate the weaknesses of the proposed C-SRIP model, we examine a few example images that result in the worst SR results according to the SSIM score in Fig.~\ref{failure_cases}. We identify a few potential reasons for the poor SR performance, i.e.:

\begin{itemize}
	\item \textit{High-frequency details}. Images \ref{failure_cases}a, \ref{failure_cases}b and \ref{failure_cases}d contain a great amount of high-frequency details (background, hair). Our SR network is guided by the face-recognition models that focus on the face and ignore other regions. %, which results in lower scores.
    \item \textit{Significant occlusion}. In images \ref{failure_cases}a and \ref{failure_cases}f, the face is partially occluded by a foreground object. The occlusion changes the global facial appearance, which adversely affects the reconstruction capabilities of C-SRIP.
    \item \textit{Significant pose variations.} In images \ref{failure_cases}e, the subject's face is partially obscured due to the profile pose. Few samples in our training dataset feature profile poses, which deteriorates performance on this type of facial images.
    \item \textit{Low-quality HR image.} Image \ref{failure_cases}c has a significant amount of noise, which is reduced during down-sampling and cannot be reconstructed. %The quality of the face reconstruction is consequently also reduced when measured through SSIM (or PSNR) values.
\end{itemize}

\vspace{-3mm}
\section{Conclusion}\label{Sec: conclusion}

We have presented a novel CNN-based model for identity-preserving face hallucination from very low-resolution images (i.e., $24\times 24$ pixels) at high magnification factors. We have shown that the proposed model improves SR results on face images, compared to both existing general super-resolution and face hallucination models. In terms of future work, we see the possibility of adapting our model to other modalities, e.g. to video sequences via recurrent attention models.

\bibliographystyle{splncs}
\bibliography{egbib}

\clearpage

\appendix{}

\section{Appendix}

In this section we present some additional results to further highlight the merits of our C-SRIP model. Similarly to the main paper, we use images from the LFW dataset~\cite{huang2007labeled} (down-sampled by smoothing the original HR images followed by sub-sampling) as our test data. All inputs to the C-SRIP model are of size $24\times 24$ pixels. %Specifically, we present some results for lower magnification factors and  

\subsection{Results for small magnification factors}

\begin{figure}[h!]
\centering
\includegraphics[width=0.85\columnwidth]{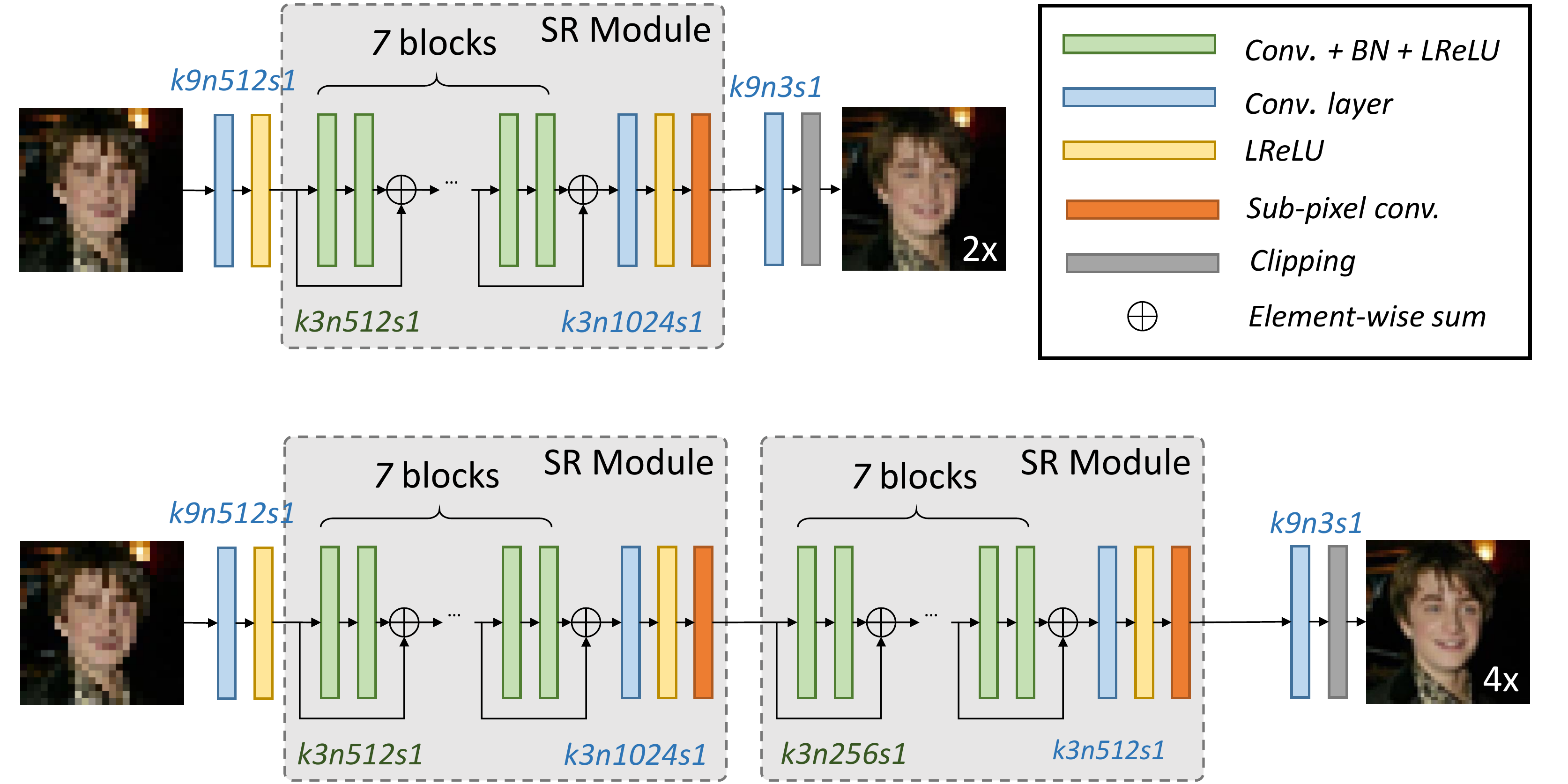}
\caption{Illustration of the intermediate results generated with the SR modules of the C-SRIP model. The top row shows the output at a magnification factor of $2\times$ and the bottom row shows the output at $4\times$. We again use the $kXnXsX$ notation introduced in~\cite{ledig2017photo} to denote convolutional layers with $n$ filters of size $k\times k$, applied with stride $s$.}
\label{fig: 2x4x}
\end{figure}

\begin{figure}[h!]
\centering
\begin{minipage}{0.495\textwidth}
\includegraphics[width=1\textwidth]{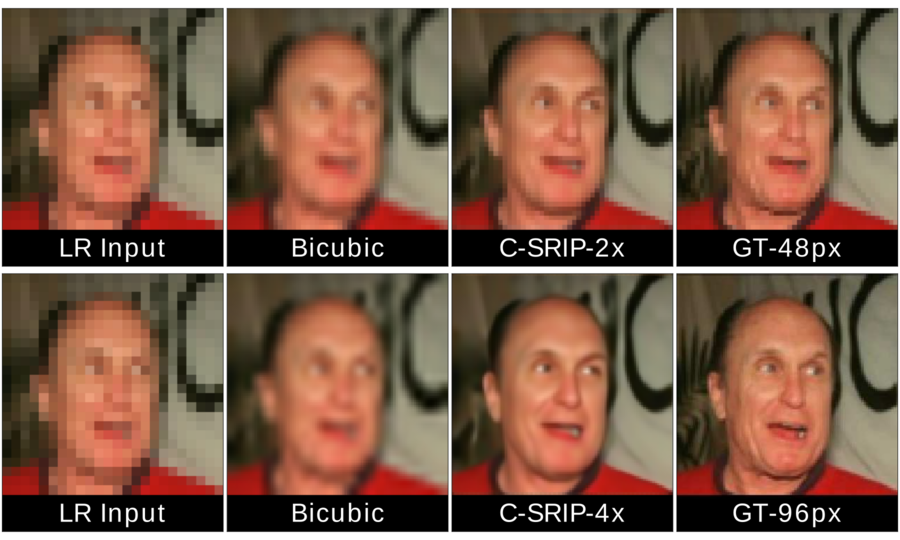}%,trim=1mm 0mm 0mm 0mm, clip
\end{minipage}
\hfill
\begin{minipage}{0.495\textwidth}
\includegraphics[width=1\textwidth]{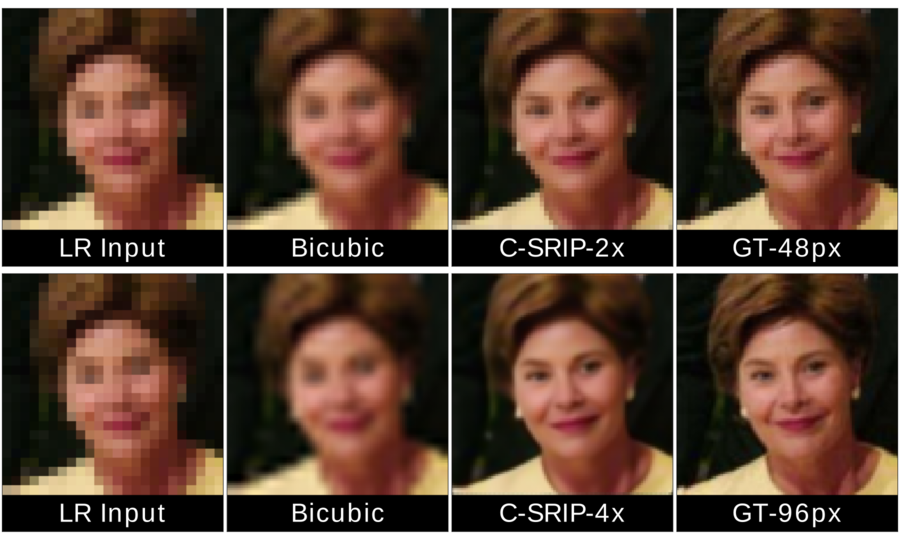}%,trim=1mm 0mm 0mm 0mm, clip
\end{minipage}
\begin{minipage}{0.495\textwidth}
\includegraphics[width=1\textwidth]{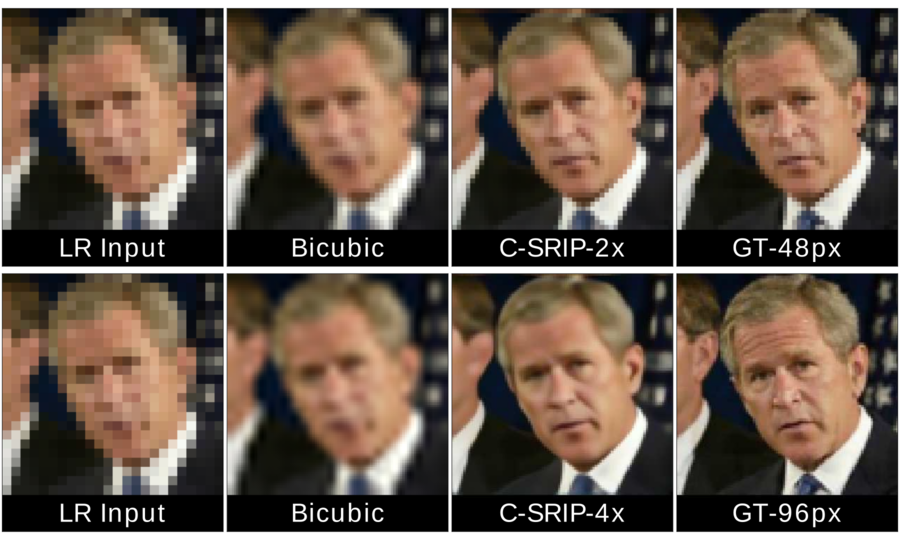}%,trim=1mm 0mm 0mm 0mm, clip
\end{minipage}
\hfill
\begin{minipage}{0.495\textwidth}
\includegraphics[width=1\textwidth]{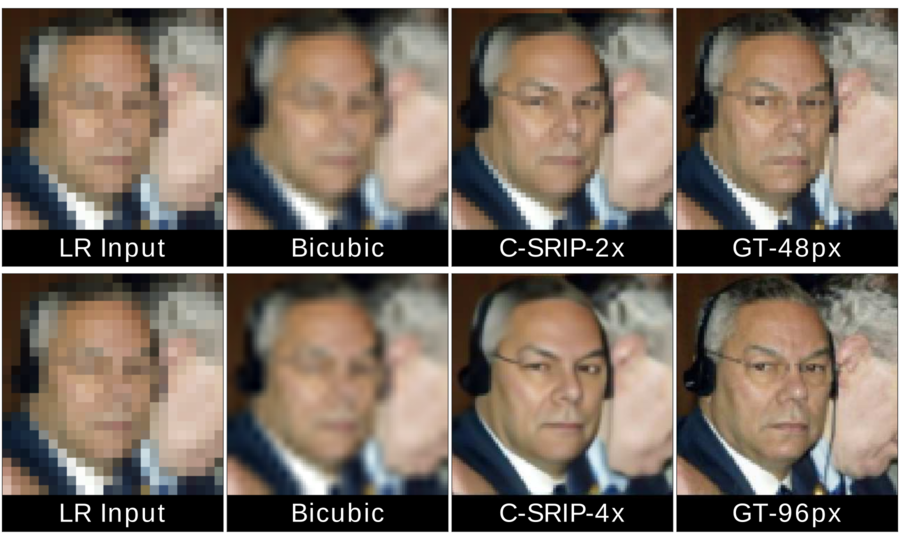}%,trim=1mm 0mm 0mm 0mm, clip
\end{minipage}
\begin{minipage}{0.495\textwidth}
\includegraphics[width=1\textwidth]{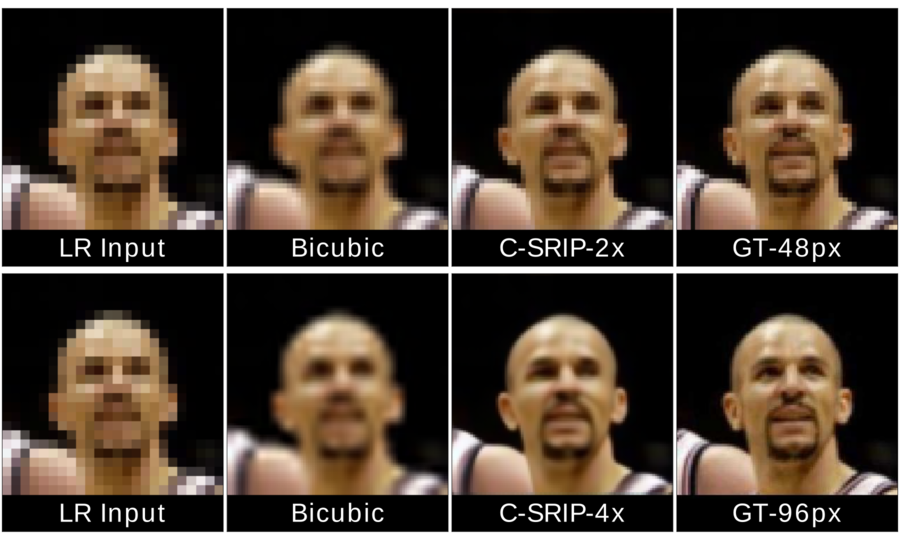}%,trim=1mm 0mm 0mm 0mm, clip
\end{minipage}
\hfill
\begin{minipage}{0.495\textwidth}
\includegraphics[width=1\textwidth]{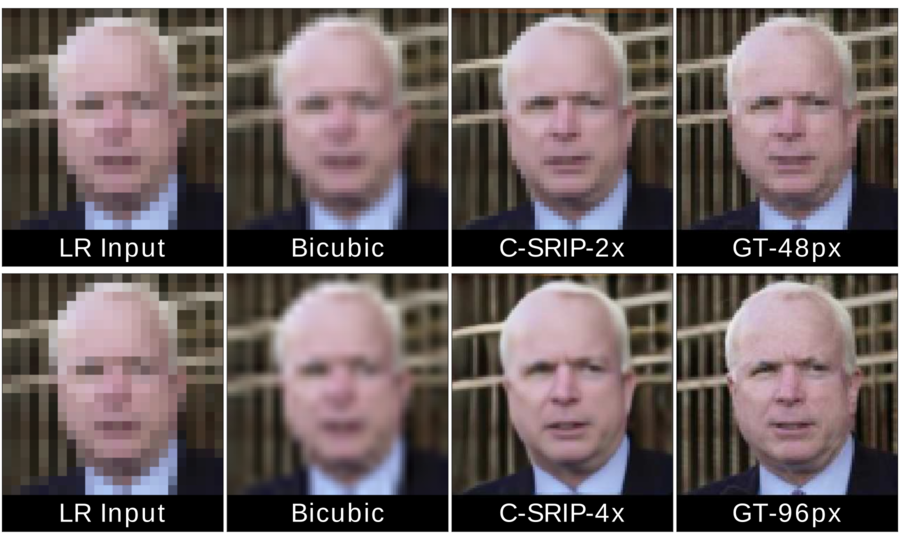}%,trim=1mm 0mm 0mm 0mm, clip
\end{minipage}
\begin{minipage}{0.495\textwidth}
\includegraphics[width=1\textwidth]{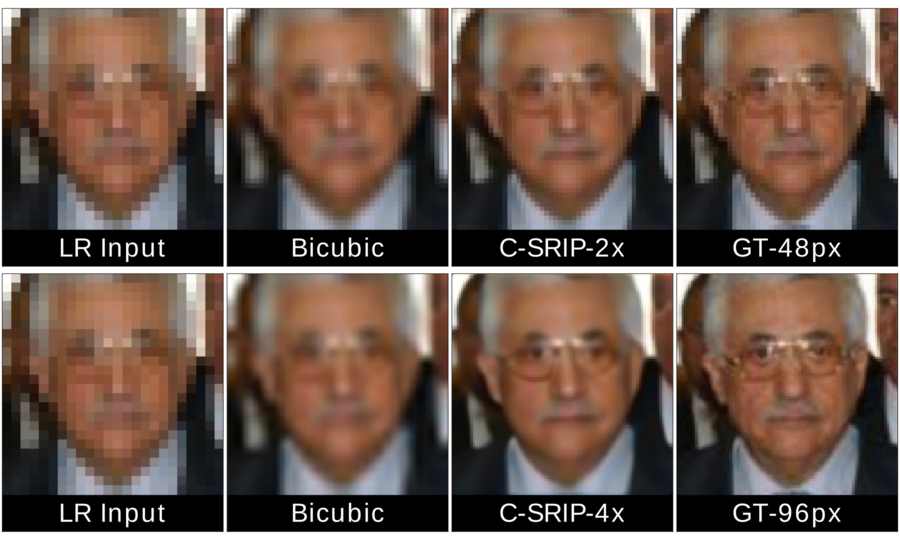}%,trim=1mm 0mm 0mm 0mm, clip
\end{minipage}
\hfill
\begin{minipage}{0.495\textwidth}
\includegraphics[width=1\textwidth]{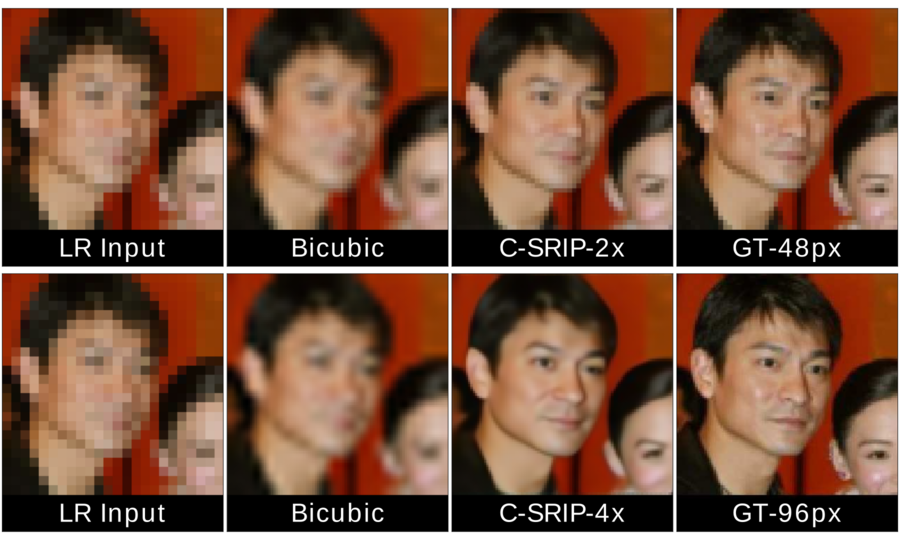}%,trim=1mm 0mm 0mm 0mm, clip
\end{minipage}\vspace{-1mm}
\caption{Qualitative results for the intermediate scales generated by our C-SRIP model. The columns correspond to (from left to right): the $24\times 24$ input image, bicubic interpolation, our results and the ground truth (GT) at either $48\times 48$ or $96\times 96$px. }\vspace{-4mm}
\label{fig: intermediate}
\end{figure}

We first demonstrate the performance of the C-SRIP model for lower magnification factors, i.e., $2\times$ and $4\times$, that produce images of size $48\times 48$ pixels and $96\times 96$ pixels, respectively, given $24\times 24$ pixel LR inputs. These images correspond to the intermediate results of the C-SRIP model that were not used for the experiments in the main part of the paper and are generated by the first and second SR module of C-SRIP as shown in Fig.~\ref{fig: 2x4x}. A few illustrative SR examples generated at $2\times$ and $4\times$ the input scale are presented in Fig.~\ref{fig: intermediate}.

%Same example results for the intermediate scales are presented in Fig.~\ref{}

We observe that  our model achieves realistic SR results even for small magnification factors. That is, even when the images are upscaled to a (still modest) size of $48\times 48$ or $96\times 96$ pixels, the hallucinated images preserve the identity of the subjects reasonably well, despite the limited performance of the SqueezNet models at these scales and, consequently, the relatively weak identity constraint applied during training. It needs to be noted that none of the presented subjects has been included in our training data.  

\subsection{Improving the visual quality of the hallucinated images}

\begin{figure}[h!]
\centering
\begin{minipage}{0.495\textwidth}
\includegraphics[width=1\textwidth]{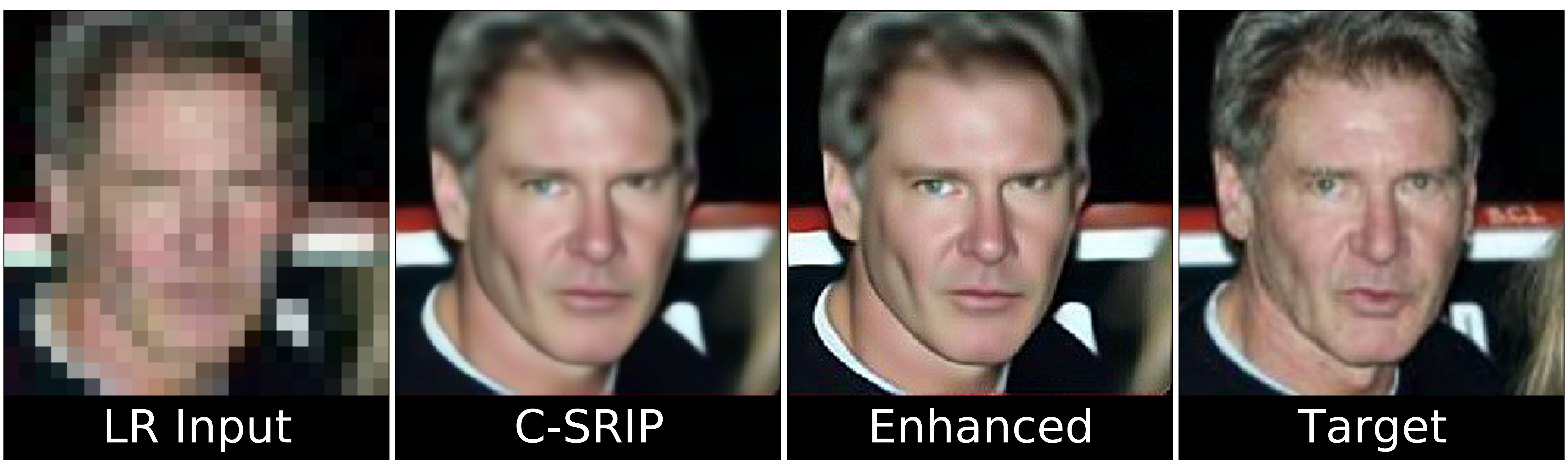}%,trim=1mm 0mm 0mm 0mm, clip \epsfig{file=foo,a=b,x=y}
\end{minipage}
\hfill
\begin{minipage}{0.495\textwidth}
\includegraphics[width=1\textwidth]{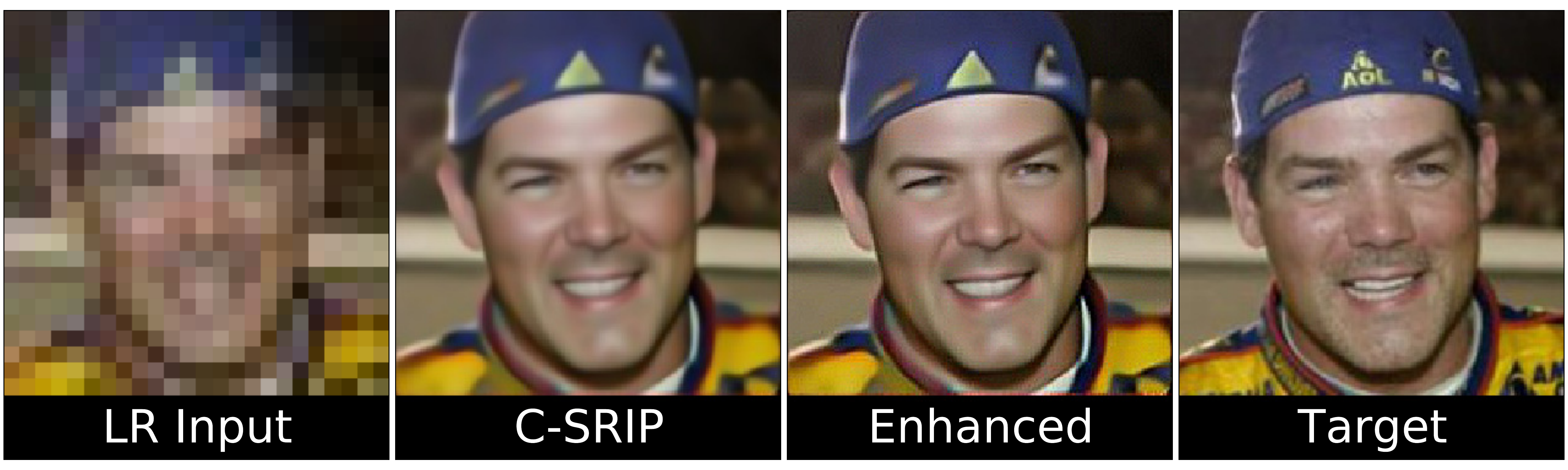}%,trim=1mm 0mm 0mm 0mm, clip
\end{minipage}
\begin{minipage}{0.495\textwidth}
\includegraphics[width=1\textwidth]{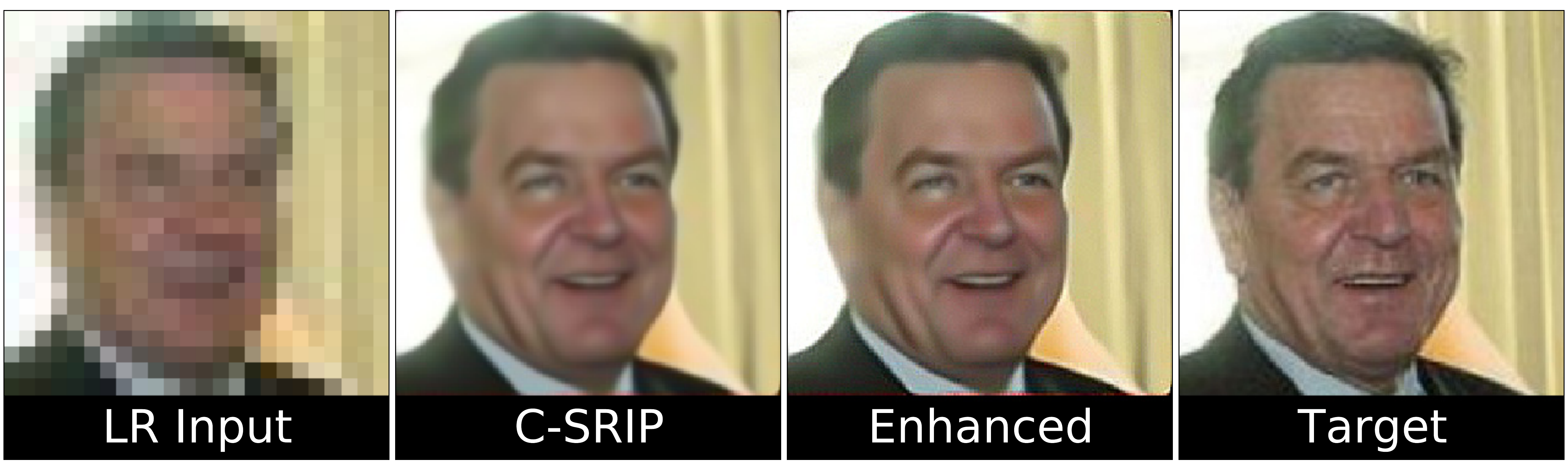}%,trim=1mm 0mm 0mm 0mm, clip
\end{minipage}
\hfill
\begin{minipage}{0.495\textwidth}
\includegraphics[width=1\textwidth]{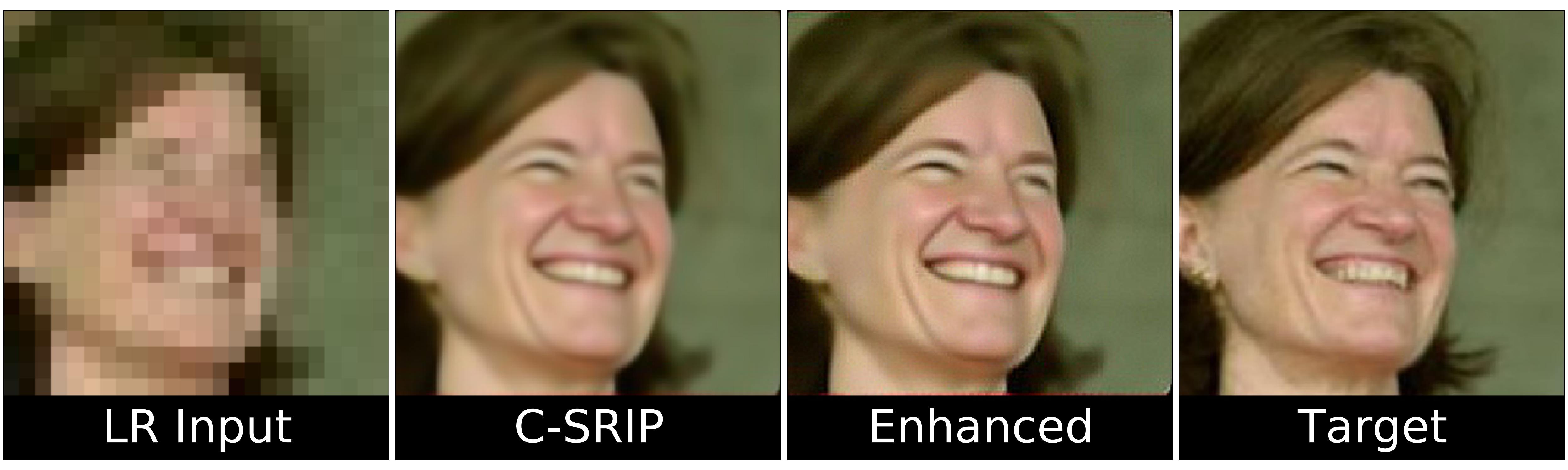}%,trim=1mm 0mm 0mm 0mm, clip
\end{minipage}
\begin{minipage}{0.495\textwidth}
\includegraphics[width=1\textwidth]{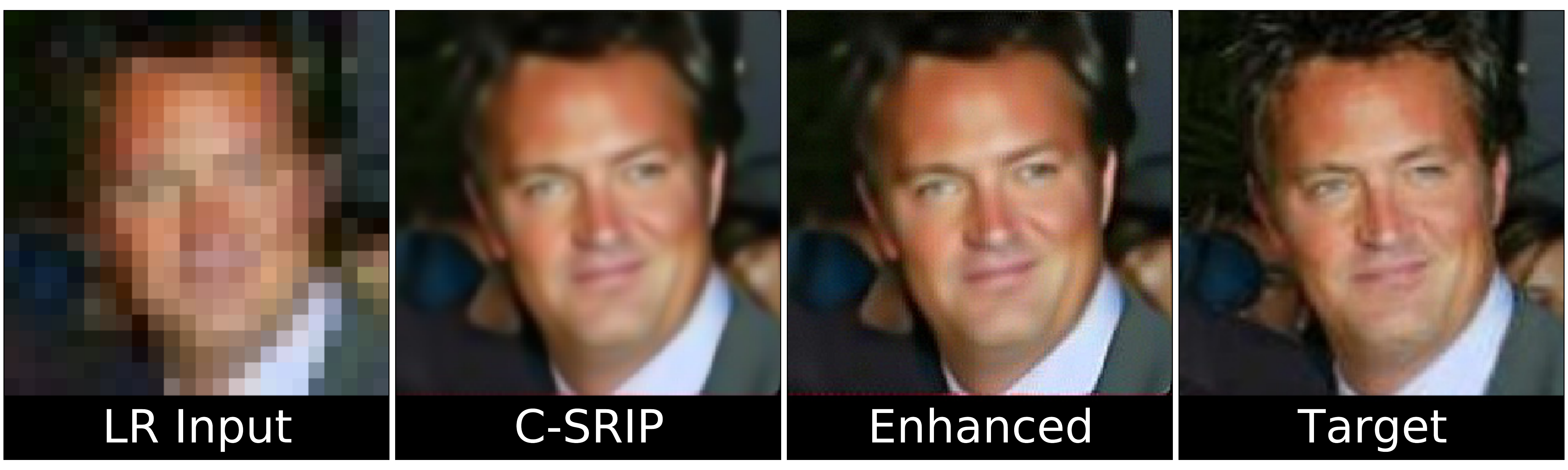}%,trim=1mm 0mm 0mm 0mm, clip
\end{minipage}
\hfill
\begin{minipage}{0.495\textwidth}
\includegraphics[width=1\textwidth]{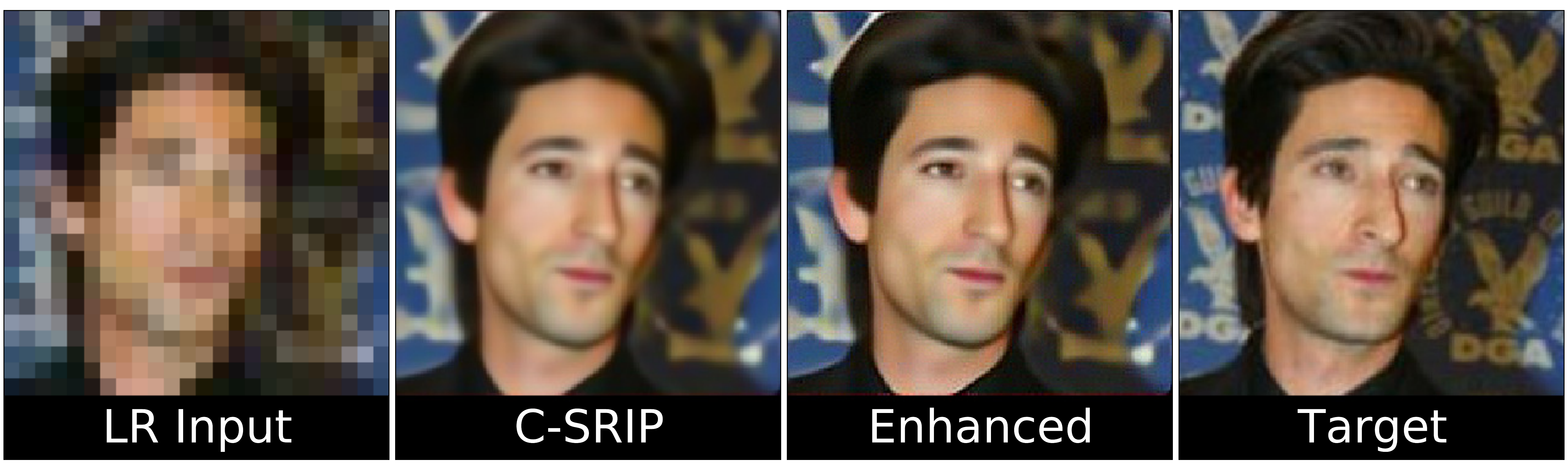}%,trim=1mm 0mm 0mm 0mm, clip
\end{minipage}
\caption{Qualitative results for SR outputs post-processed with a standard image enhancement technique (i.e., with a sharpening filter). For each $24\times 24$ LR input image (on the far left of each quadruplet) the following columns correspond to (from left to right): C-SRIP, C-SRIP with image enhancement, and the target HR image. Best viewed in high resolution.}
\label{fig: enhanced}
\end{figure}

\begin{figure}[h!]
{\centering
\includegraphics[width=1\textwidth]{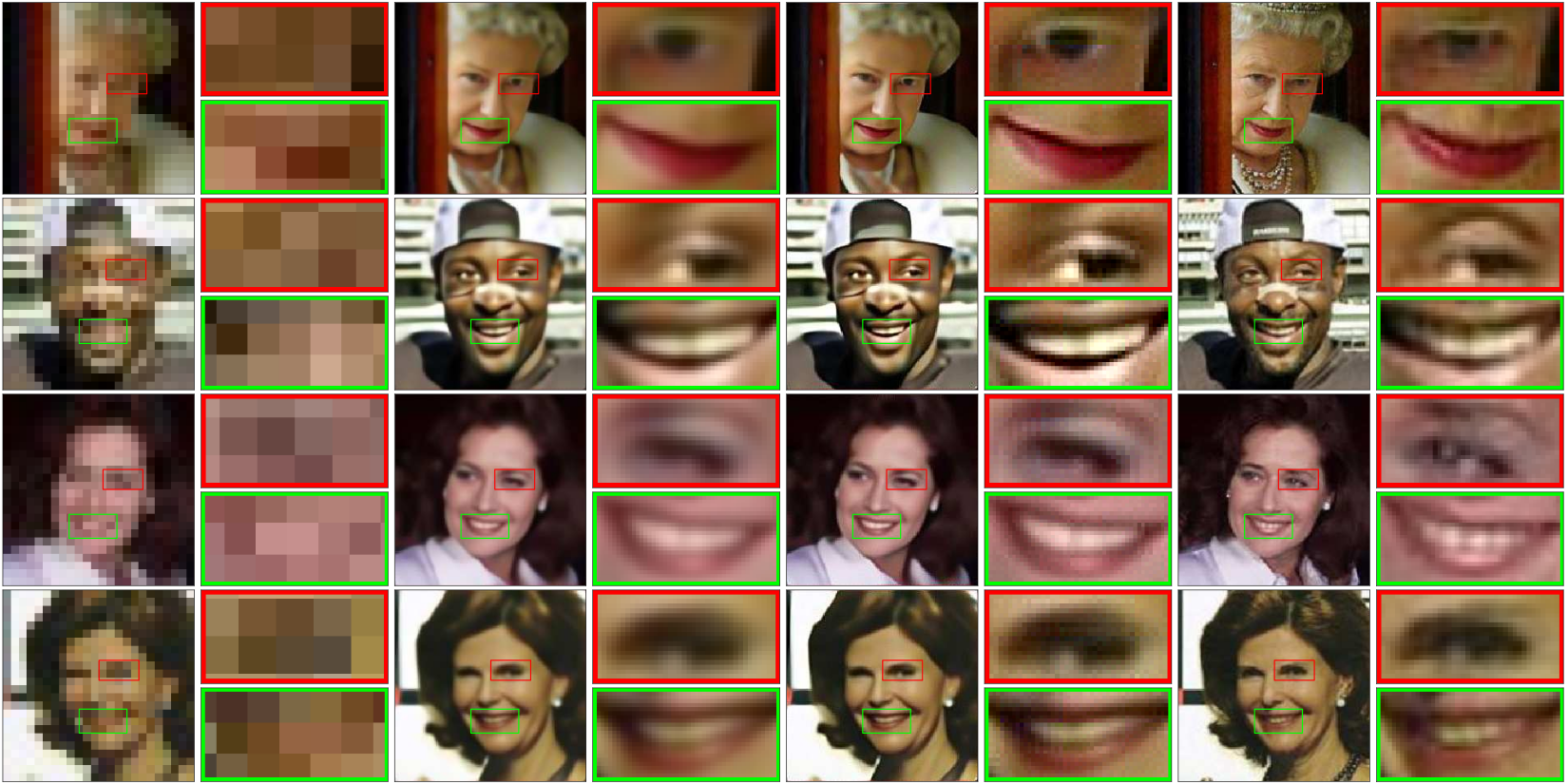}}\\%,trim=1mm 0mm 0mm 0mm, clip
\text{\scriptsize \hspace{12mm}LR Input \hspace{16mm}C-SRIP \hspace{14mm}C-SRIP Enhanced \hspace{14mm}HR Target}
\caption{Qualitative results for SR outputs post-processed with a standard image enhancement technique (i.e., with a sharpening filter) with highlighted image details. For each $24\times 24$ LR input image (on the far left of each quadruplet) the following columns correspond to (from left to right): C-SRIP, C-SRIP with image enhancement and the target HR image. Best viewed in high resolution.}
\label{fig: details}
\end{figure}

It is possible to further improve on the (perceived) visual quality of the SR images produced by the C-SRIP model (for large magnification factors of $8 \times$) by utilizing simple image enhancement techniques. In Fig.~\ref{fig: enhanced} and Fig.~\ref{fig: details} we show some examples, where a standard $3\times 3$ sharpening filter (i.e., $\left[0, -1, 0;-1,5,-1;0, -1, 0 \right]$) is applied on the SR outputs to amplify the high frequency components of the generated images. The result of applying such post-processing steps are significantly sharper in crisper SR images. However, in terms of summary statistics (i.e., average SSIM and PSNR scores) these are not competitive to the results reported in the main part of the paper - the sharpening operation deteriorates (quantitatively measured) performance. In Fig.~\ref{fig: enhanced} and Fig.~\ref{fig: details} we also include results for some examples that were already presented in the main part of the paper to facilitate implicit comparisons with competing methods. 

Interestingly, after the post-processing some of the SR images  appear sharper than the original HR targets. This can be partially explained by the presence of noise in the target images that is not present in the SR reconstructions and the higher image contrast after enhancement that contributes towards the perception of higher-quality images. %  to the presence of noise

\subsection{Quantitative results on the impact of the SSIM loss}
\begin{table}[!tb]
\centering
\caption{PSNR and SSIM scores obtained on the training data with the MSE- and SSIM-based losses.}\label{tab: simple_generators}
\small
\label{tab: training_chars}
\begin{tabular}{|l|r|r|r|r|r|} \hline
 & MSE-based loss & SSIM-based loss \\ \hline \hline
PNSR $[dB]$   & $28.3275$ & $29.0227$ \\  \hline
SSIM  & $0.9189$ & $0.9325$ \\  \hline
\end{tabular}
\end{table}

\begin{table}[tb]
\centering
\caption{Comparison of the PSNR and SSIM scores on the test data obtained with the MSE- and SSIM-based losses.}
\small
\footnotesize
\label{ablation_study}
\begin{tabular}{|l|r|r|r|r|r|} \hline
 & MSE-based loss & SSIM-based loss \\ \hline \hline
 PSNR $[dB]$ & $26.1748$ & $26.0251$ \\ \hline
 SSIM & $0.7547$ & $0.7579$\\ \hline
\end{tabular}
\end{table}

Next, we present some (additional) quantitative results related to the proposed SSIM loss. Our SSIM formulation uses convolutions with a discrete Gaussian kernel, $\mathbf{g}$ - see Eq. (3), to approximate the local averages used with the original SSIM and is, therefore, easily implementable using standard deep learning frameworks. As emphasized in the main part of the paper, the result of using the proposed SSIM-based loss are significantly better training characteristics in terms of faster convergence and lower PSNR and SSIM scores on the training data as shown in Table~\ref{tab: training_chars}. Here, the results are presented for the simplest architecture from the ablation study (Section 4.3), where \textit{i)} the images are processed through a series of $21$ residual blocks, \textit{ii)} all three upscaling layers are placed at the end of the SR network, and \textit{iii)} supervision is applied only at the output of the model.  

The proposed SSIM-based loss ensures significantly better performance scores during training. Even though the MSE-based loss is directly proportional to the PSNR score, our SSIM-based loss results in a lower average PSNR score on the training data, which suggests that a better optimum was found by the backpropagation-based learning procedure. On the test data the proposed loss still improves on the SSIM score, but offers no improvements in terms of PSNR value as shown in Table~\ref{ablation_study} - this is already highlighted  in  the ablation study of the main part of the paper.

\end{document}